%% file: neurips_2025.tex
\documentclass{article}

\usepackage[preprint]{neurips_2025}

\usepackage[utf8]{inputenc} %
\usepackage[T1]{fontenc}    %
\usepackage{hyperref}       %
\usepackage{url}            %
\usepackage{booktabs}       %
\usepackage{amsfonts}       %
\usepackage{nicefrac}       %
\usepackage{microtype}      %
\usepackage{xcolor}         %

\usepackage{amsmath}
\usepackage{amssymb}
\usepackage{mathtools}
\usepackage{amsthm}

\usepackage{utils}

\usepackage{booktabs}
\usepackage{multirow}
\usepackage{adjustbox}
\usepackage{makecell}

\usepackage{subcaption}
\usepackage{graphicx}
\usepackage{wrapfig}
\usepackage{url}
\definecolor{custompink}{RGB}{255,20,147}
\hypersetup{
    colorlinks = true,
    urlcolor = custompink
}

\title{\vspace{-2.5mm}\raisebox{-1.8mm}{\includegraphics[scale=0.04]{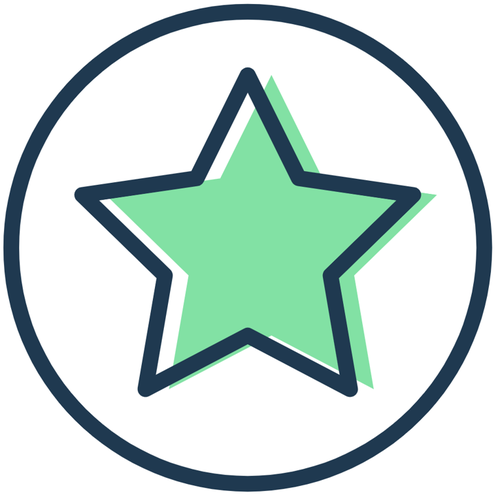}} STAR-R1: Spatial TrAnsformation Reasoning by Reinforcing Multimodal LLMs}

\author{%
  Zongzhao Li$^{1}$\thanks{Equal contribution.} \quad
  Zongyang Ma$^{2}$\footnotemark[1] \quad
  Mingze Li$^1$\quad
  Songyou Li$^1$ \quad
  Yu Rong$^{3,4}$ \\ 
  \textbf{Tingyang Xu}$^{3,4}$ \quad
  \textbf{Ziqi Zhang}$^2$ \quad
  \textbf{Deli Zhao}$^{3,4}$ \quad
  \textbf{Wenbing Huang}$^{1}$\thanks{Corresponding author.}\\ 
  $^1$Gaoling School of Artificial Intelligence, Renmin University of China \\ $^2$MAIS, Institute of Automation, Chinese Academy of Sciences \\ $^3$DAMO Academy, Alibaba Group, Hangzhou,
China \ $^4$Hupan Lab, Hangzhou, China  \\
  \texttt{lizongzhao2023@ruc\!.\!edu\!.\!cn}, \ \ 
\texttt{mazongyang2020@ia\!.\!ac\!.\!cn} \\
\texttt{hwenbing@126\!.\!com} \\
}

\begin{document}

\maketitle

\vspace{-0.5cm}
\begin{abstract}
  Multimodal Large Language Models (MLLMs) have demonstrated remarkable capabilities across diverse tasks, yet they lag significantly behind humans in spatial reasoning. We investigate this gap through Transformation-Driven Visual Reasoning (TVR), a challenging task requiring identification of object transformations across images under varying viewpoints. While traditional Supervised Fine-Tuning (SFT) fails to generate coherent reasoning paths in cross-view settings, sparse-reward Reinforcement Learning (RL) suffers from inefficient exploration and slow convergence. To address these limitations, we propose STAR-R1, a novel framework that integrates a single-stage RL paradigm with a fine-grained reward mechanism tailored for TVR. Specifically, STAR-R1 rewards partial correctness while penalizing excessive enumeration and passive inaction, enabling efficient exploration and precise reasoning. Comprehensive evaluations demonstrate that STAR-R1 achieves state-of-the-art performance across all 11 metrics, outperforming SFT by 23\% in cross-view scenarios. Further analysis reveals STAR-R1’s anthropomorphic behavior and highlights its unique ability to compare all objects for improving spatial reasoning. Our work provides critical insights in advancing the research of MLLMs and reasoning models.
  The codes, model weights, and data will be publicly available at \url{https://github.com/zongzhao23/STAR-R1}.
  
\end{abstract}

\section{Introduction}
\label{sec:introduction}

Multimodal Large Language Models (MLLMs) \cite{bai2023qwenvl, wang2024qwen2vl, bai2025qwen25vl, chen2024internvl, chen2024internvl1.5, team2024internvl2, chen2024expanding2.5, yao2024minicpm, agrawal2024pixtral, Emu2, wang2024emu3, chen2025janus} have demonstrated remarkable progress in recent years, excelling at tasks such as visual question answering~\cite{fu2024mme, yu2023mm, yue2024mmmu}, text-to-image generation~\cite{ghosh2023geneval, hu2024ella}, and video understanding~\cite{li2024mvbench,zhou2024mlvu,wang2024lvbench,fu2024video,chen2024we}. Despite these advances, existing MLLMs still struggle with spatial reasoning \cite{mayer2025ivispar, tang2025lego, jia2025omnispatial, liu2025ir3d}---a fundamental aspect of human intelligence. Given this critical gap in emulating human-like spatial cognition, it becomes imperative to develop a targeted framework for assessing such aspect. To this end, this paper takes the challenging task of Transformation Driven Visual Reasoning (TVR) \cite{hong2021transformation} as a starting point to investigate MLLMs' capabilities in spatial understanding. 
By comparing two given images containing multiple objects, TVR involves a series of continuous thinking and reasoning steps to derive the correct answer, including identifying which object is transformed and what transformation is applied to this object. Moreover, objects in the two images may be presented in different viewpoints, which further complicates the task and poses difficulties for existing MLLMs.
Our experiments show that, the powerful closed-source commercial model GPT-4o \cite{hurst2024gpt} achieves only 23.5\% accuracy on this task, further demonstrating that this problem poses a significant challenge to existing MLLMs.

\begin{figure*}[]
    \centering    \includegraphics[width=1.0\linewidth]{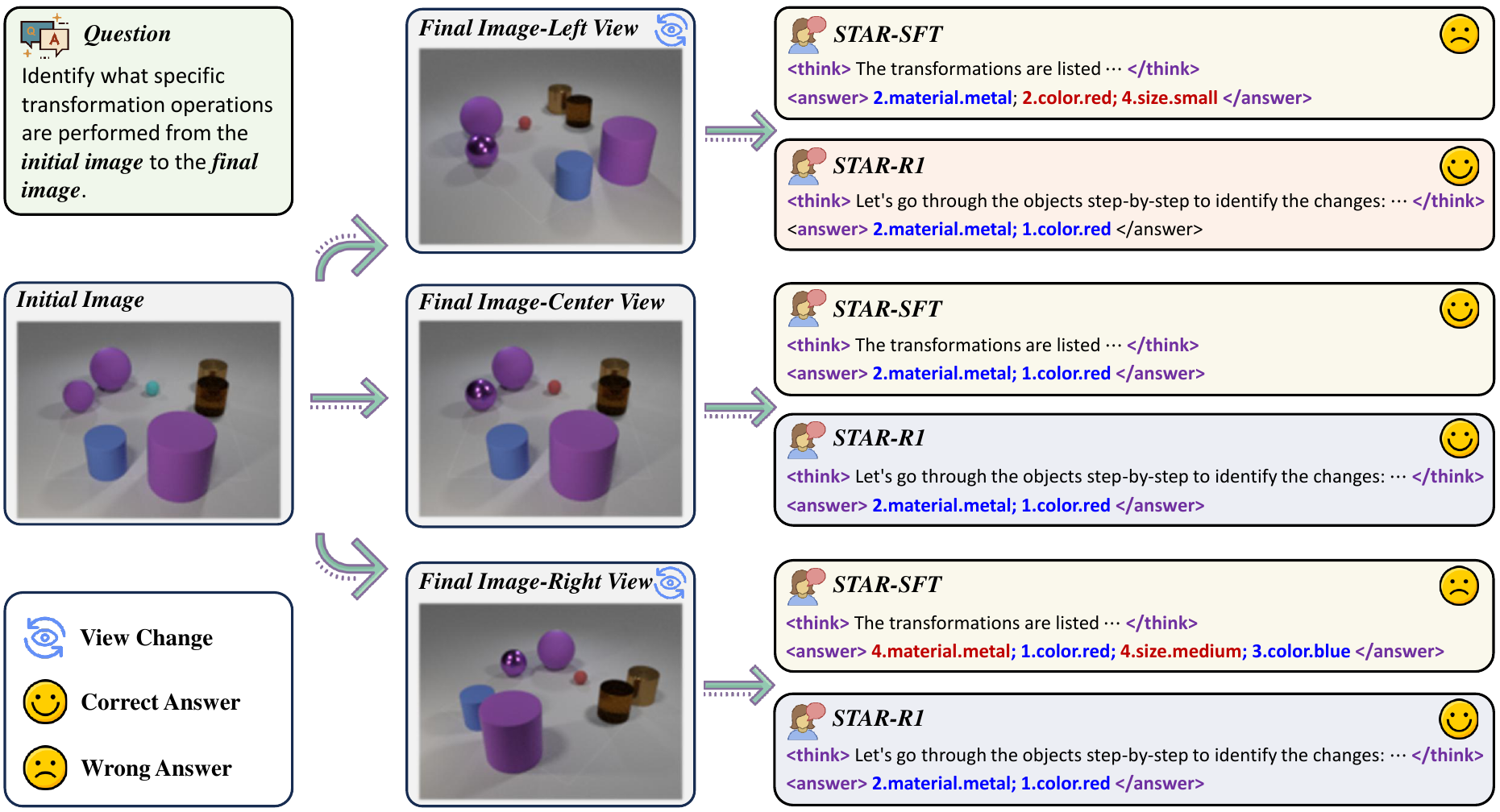}
    \caption{Comparison of addressing the TVR task by STAR-SFT and STAR-R1. STAR-SFT finetunes  MLLMs with supervised instructions, while STAR-R1 employs RL-guided thinking.}
    \label{fig:intro}
    \vspace{-15pt}
\end{figure*}

While Supervised Fine-Tuning (SFT) can improve MLLMs’ performance on TVR, this approach remains insufficient. 
Since successfully solving TVR requires step-by-step Chain-of-Thought (CoT) \cite{,wei2022chain} reasoning, merely replicating annotations of final object and attribute changes fails to yield robust performance, particularly in view-shifting scenarios. 
Recently, training LLM by Reinforcement Learning (RL), exemplified by DeepSeek-R1-Zero~\cite{guo2025deepseek}, has demonstrated significant advantages in generating the correct CoT, offering a promising solution for the TVR task. 
However, the naive GRPO method \cite{shao2024deepseekmath} in DeepSeek-R1-Zero, designed for tasks like Mathematical Problem Solving (MPS) \cite{cobbe2021training, hendrycks2021measuring, hendrycks2020measuring,zhong2023agieval,wei2023cmath}, uses sparse reward based on the binary verification signal. 
When directly adapted to the TVR task, it results in low exploration efficiency and slow convergence. 
In the MPS task, the final answer, such as the option of the multiple-choice question or the value of the fill-in-the-blank question, has only two possibilities: correct or incorrect, without intermediate state of partial correctness. 
In contrast, the final answer in the TVR task is a sequence composed of multiple objects with multiple transformations. During early training, the model often produces partially correct responses; for instance, it may correctly identify the object whose attribute has changed but fail to specify the exact transformation applied. Under a binary reward setting, the model rarely receives positive feedback, hindering its ability to determine effective exploration directions.

To address these challenges, we propose to integrate dense reward plus punishment into GRPO, thereby constructing a new method dubbed Spatial TrAnsformation Reasoning with R1 (STAR-R1).
STAR-R1 is built upon two core designs. 
On the one hand, it introduces dense rewards by accounting for partial answer correctness, progressively assigning fine-grained rewards to three-level answers. 
Specifically, for the first-level answer, the model identifies the objects with modified attributes, receiving a minimal reward. For the second-level answer, it correctly predicts the changed attributes, earning a medium reward. For the third-level answer, the model accurately identifies the full object-attribute-transformation triplet, receiving the highest reward. On the other hand, STAR-R1 penalizes the model for either excessively enumerating object-attribute-change triplets to maximize rewards or remaining inactive to avoid incorrect predictions, ensuring active exploration without shortcuts. This mechanism design instills into the model the learning philosophy that prioritizes attempting potentially incorrect answers over overlooking possible correct ones.

Through rigorous data filtering and sampling pipeline, we select training/test samples from the original TRANCE dataset~\citep{hong2021transformation}. To comprehensively evaluate existing MLLMs on the TVR task, our test set includes samples whose initial and final images are captured either from the same view (In-Domain) or from more challenging different views (Out-of-Domain), totaling 4.5K samples. Extensive evaluations on both closed-source commercial and open-source MLLMs demonstrate that the proposed STAR-R1 not only significantly outperforms other models across all types of attribute transformations, but also exceeds the performance of SFT methods under the same training settings. Notably, in Out-of-Domain (OOD) tasks, STAR-R1 achieves a substantial 23\% performance advantage over SFT-based approaches. Furthermore, through in-depth analysis of experimental results, we reveal STAR-R1’s \textbf{anthropomorphic behavior} and identify the core reason for RL’s superiority: it majorly compares all objects between the initial and final scenes, thereby improving reasoning in OOD scenarios. We also observe an intriguing phenomenon where the model’s response length initially decreases and then increases during training, indicating the specificity of the TVR task compared to other common reasoning tasks. 
Our contributions are summarized as follows:

\begin{itemize}[left=1.0em, itemsep=2pt, parsep=2pt, topsep=0pt]
    \item We propose STAR-R1, a multimodal reasoning model trained via a \textbf{single-stage pure RL} paradigm akin to DeepSeek-R1-Zero. By integrating a customized reward function for RL optimization, STAR-R1 exhibits superior reasoning capabilities in spatial understanding.

    \item In the TVR task, STAR-R1 achieves \textbf{the best} performance across all eleven evaluation metrics. Moreover, in the cross-view OOD task setting, it delivers a \textbf{23\%} TAcc gain over SFT, validating the R1-Zero paradigm’s potential in unlocking reasoning capabilities of the TVR task.

    \item Through in-depth analysis of experimental results, we uncover STAR-R1’s \textbf{anthropomorphic behavior} and elucidate the mechanistic advantages of RL over SFT. We also systematically investigate the dynamic changes in response length. These findings provide unique insights for advancing MLLMs in spatial reasoning.
\end{itemize}

\section{Related Work}

\noindent \textbf{Multimodal Large Language Model.}
MLLMs have made rapid advancements recently, achieving significant improvements in various image \cite{fu2024mme, mathew2021docvqa} and video understanding \cite{li2024mvbench, mangalam2023egoschema} tasks. 
The early LLaVA series models \cite{liu2023llava, liu2024llava-1.5, li2024llava-next, li2024llava-onevision} are typical representatives of MLLMs, which use a simple MLP as the projectioner to map the CLIP-encoded \cite{radford2021learning, sun2023eva-clip} visual features into the semantic space of Large Language Models (LLMs) \cite{touvron2023llama, touvron2023llama2, grattafiori2024llama3, bai2023qwen, yang2024qwen2.5}, enabling the visual-language joint understanding. 
This architecture is subsequently widely adopted by more powerful models such as the QwenVL series \cite{bai2023qwenvl, wang2024qwen2vl, bai2025qwen25vl} and InternVL series \cite{chen2024internvl, chen2024internvl1.5, team2024internvl2}, significantly narrowing the performance gap between open-source MLLMs and commercial-grade closed-source benchmark models like GPT-4o \cite{hurst2024gpt} and Gemini series \cite{team2024gemini}.
Despite these advancements, current MLLMs still struggle with complex reasoning tasks, especially those involving spatial reasoning \cite{mayer2025ivispar, tang2025lego}. 
To this end, we treat the TVR task as a pivotal point, concentrating on exploring effective methods for enhancing the spatial reasoning capabilities of MLLMs.

\noindent \textbf{Multimodal Reasoning Model.}
Research on the reasoning capabilities of MLLMs has attracted significant attention in recent years. Early studies primarily employed Chain-of-Thought (CoT) fine-tuning to enhance model reasoning abilities~\citep{xu2024llava, xu2025redstar}, while the recent success of DeepSeek-R1~\citep{guo2025deepseek} has promoted the application of rule-based Reinforcement Learning in this field~\citep{chen2025r1v, shen2025vlm, yang2025deepcritic, fan2025sophiavl, liang2025swsselfawareweaknessdrivenproblem, li2025temporal, wang2025solidgeomeasuringmultimodalspatial, yang2025towards, song2025maniplvmr1reinforcementlearningreasoning}. Notably, MM-Eureka~\citep{meng2025mm} adopts an online filtering strategy and a two-stage multimodal RL training framework to achieve significant improvements in mathematical reasoning tasks; LMM-R1~\citep{peng2025lmm}, which enhances multimodal reasoning capabilities through a two-phase training paradigm combining text-only RL and multimodal RL; and Video-R1~\citep{feng2025video}, which achieves breakthroughs in video reasoning tasks via a two-stage approach involving supervised fine-tuning (SFT) followed by RL training. However, existing methods exhibit several limitations: (1) most rely on naive GRPO strategies, (2) they predominantly adopt two-stage training paradigms, and (3) there is a lack of task-specific customization and in-depth exploration of pure RL training (e.g., like DeepSeek-R1-Zero). To address these gaps, we introduces a tailored rule-based RL strategy for the TVR task and proposes a single-stage, pure RL training paradigm. Our approach not only significantly enhances multimodal reasoning performance but also provides novel theoretical insights into RL training mechanisms.

\section{Our Method: STAR-R1}
\label{sec:method}

\begin{figure*}[]
    \centering    \includegraphics[width=1.0\linewidth]{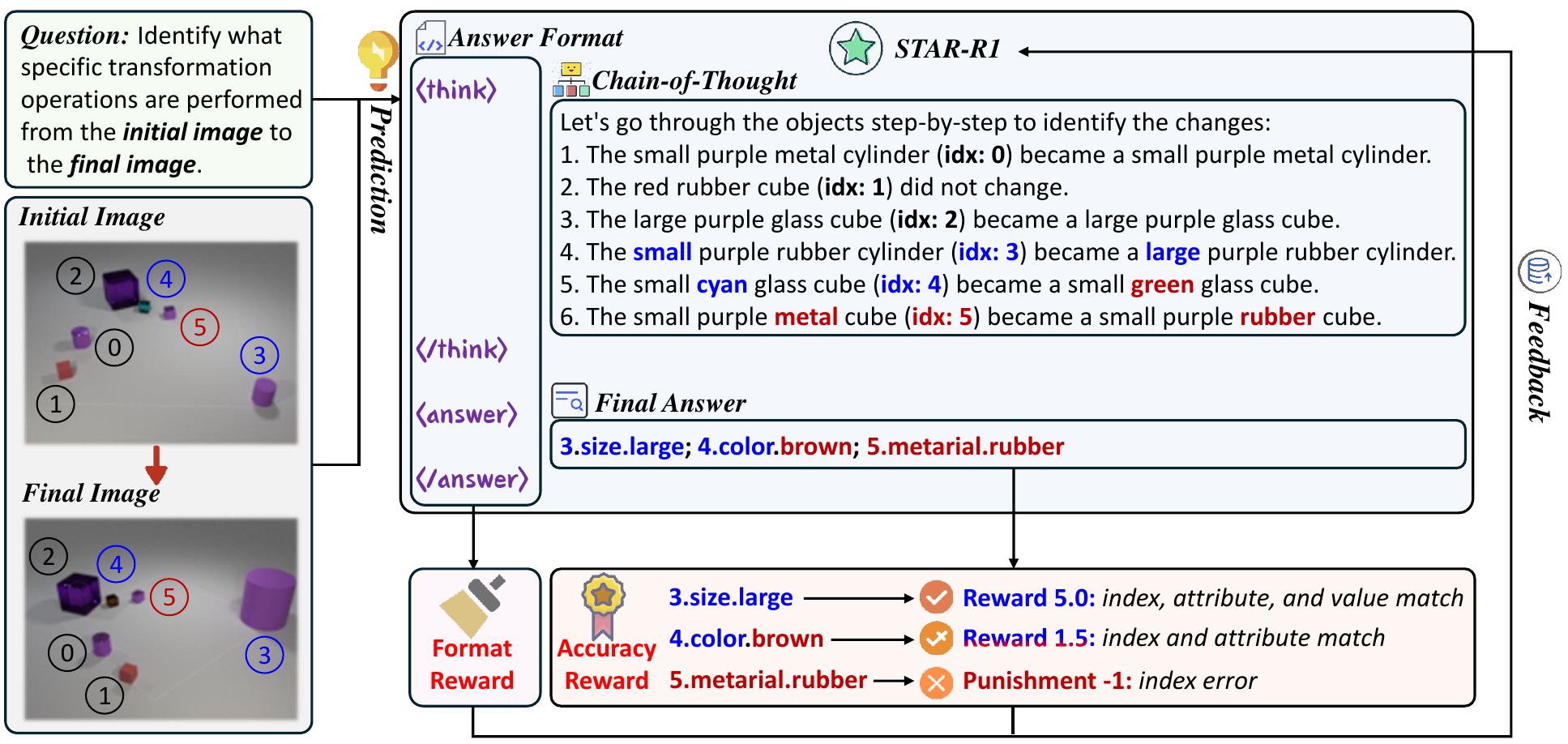}
    \caption{The STAR-R1 Framework. Given initial/final images and a question, STAR-R1 establishes object correspondences by systematically comparing all objects across both images. It generates Step-by-step reasoning in \textcolor{violet!85!black}{<think></think>} tags before delivering the final answer in \textcolor{violet!85!black}{<answer></answer>} tags. We demonstrate the reward function of STAR-R1 using a partially correct case study. Correct/incorrect answers are shown in \textcolor{blue}{blue} and \textcolor{red}{red} respectively.}
    \label{fig:model}
    \vspace{-15pt}
\end{figure*}

\subsection{Task Formulation}
\label{sec:task-form}
As illustrated in \cref{fig:model}, the TVR task is formally defined in this way: given an initial image $I^{\text{initial}}$, a final image $I^{\text{final}}$, and a question $Q$, the model $\phi$ is required to generate a sequence of transformations $T$ as the output answer. Formally, this generation process can be expressed as:
\begin{equation}\label{eq:task_dynamic}
\begin{aligned}
    T = \phi (I^{\text{initial}}, I^{\text{final}}, Q). \\
\end{aligned}
\end{equation}

Here, $T = \{t_1, t_2, \dots, t_n\}$ denotes a sequence of transformations. To avoid the enormous description space introduced by pixel-level transformations, we adopt the methodology from the previous work~\citep{hong2021transformation}, defining a single-attribute transformation of a single object as an atomic operation. In form, $t_i = (\text{index}_i, \text{attribute}_i, \text{value}_i)$, which indicates modifying $\text{attribute}_i$ to $\text{value}_i$, of the object with $\text{index}_i$. Given the multi-step nature of transformations, MLLMs must meticulously analyze fine-grained differences between the initial and final images and perform step-by-step reasoning to derive the correct transformation sequence.

\subsection{The Design of Reward Function}
\label{sec:reward}
Reward function, serves as direct feedback from the environment and  plays a pivotal role in reinforcement learning. A sparse reward space can lead to inefficient exploration by the model, often requiring extended training periods to discover optimal decision paths. An excessively dense reward, on the contrary, may induce the model to adopt shortcut strategies for reward hacking. To facilitate active and efficient exploration of the solution space, our STAR-R1 integrates dense reward and punishment mechanisms into the original GRPO framework~\citep{shao2024deepseekmath}. Particularly, we incorporate not only format rewards for structural constraints but also dense reward and punishment mechanisms to provide precise reward signals. The design principles of these two reward components are as follows.

\textbf{Format Reward}\quad
The format reward enforces structural constraints on model outputs, mandating that reasoning processes should appear within <\text{think}></\text{think}> tags while final answers must be enclosed in <\text{answer}></\text{answer}> tags. Compliance of this formate yields a reward $R^{\text{format}}$ of 1, whereas violations result in 0.

\textbf{Accuracy Reward}\quad
In existing studies, Reinforcement Learning with Verifiable Rewards (RLVR) primarily employs binary verification signals (correct or incorrect), which can be easily computed by simple rule-based verifiers. This paradigm has been validated in multiple studies and has demonstrated promising results across various scenarios ~\citep{xie2025logic, meng2025mm}. However, in the TVR task, directly applying binary verification signals---the model only receives a reward if all transformations are correct---would lead to inefficient exploration and suboptimal model performance. Therefore, our model STAR-R1 first refines the reward granularity to each individual transformation and then decomposes the reward into each component of the transformation. Additionally, it carries out punishment mechanisms to guide the model toward correct and proactive exploration, ensuring both efficient exploration and high prediction accuracy.

For any transformation $t_i = (\text{index}_i, \text{attribute}_i, \text{value}_i)$ in the model's output set $T = \{t_1, t_2, \dots, t_n\}$, if its object index, attribute, and value exactly match a corresponding $\hat{t}_i$ in the ground-truth transformations $\hat{T} = \{\hat{t}_1, \hat{t}_2, \dots, \hat{t}_n\}$, the model is deemed to have perfectly inferred the transformation and thus receives the highest reward ($\text{reward} = 5.0$). If only the object index and attribute match but the value is incorrect, the model demonstrates accurate \textit{identification} and \textit{association} capabilities, which are critical for forming correct reasoning chains; hence, a partial reward ($\text{reward} = 1.5$) is granted. A match solely in the object index reflects preliminary \textit{identification} ability, and a minimal reward ($\text{reward} = 0.5$) is provided to encourage iterative model improvement. The positive reward described above can be formally defined as:
\begin{align}
R^{\text{pos}}(t_i) =
\begin{cases}
5.0, & \text{if all $\text{index}_i$, $\text{attribute}_i$, and $\text{value}_i$ match;} \\
1.5, & \text{if $\text{index}_i$ and $\text{attribute}_i$ match;} \\
0.5, & \text{if only $\text{index}_i$ matches.}
\end{cases}    
\end{align}

By accumulating the rewards of all  $t_i$s, we obtain $R^{\text{pos}} = \sum^{n}_{i=1}R^{\text{pos}}(t_i)$.

Building upon the positive reward design, we further introduce a dual punishment mechanism to effectively constrain the model's exploratory behavior. Relying solely on positive rewards may lead the model to engage in reward hacking by exhaustively enumerating all possible $(\text{index}_i, \text{attribute}_i, \text{value}_i)$ triplets. To mitigate this, after positive reward calculation, we further examine the predictions to verify whether the attribute transformations are inconsistent with the final image state. Each mistaken prediction incurs a punishment ($\text{reward} = -1.0$), yielding a total reward as  $-n_{\text{miss}}$ for $n_{\text{miss}}$ mistaken predictions. Furthermore, to encourage comprehensive exploration of all possible transformations, the system will impose a punishment ($\text{reward} = -(\hat{n} - n)$), when the number of predicted transformations $n$ is less than the ground truth $\hat{n}$. This mechanism teaches the model to prioritize attempting potentially correct answers over avoiding possible errors, thereby promoting continuous experimentation, in-depth analysis, and precise evaluation of each object's attribute changes, ultimately leading to steady improvements in accuracy. The punishment reward described above can be formally defined as:
\begin{align}
R^{\text{pun}} =
\begin{cases}
-n_{\text{mis}}-(\hat{n} - n), & \text{if $n$ < $\hat{n}$}; \\
-n_{\text{mis}}, & \text{otherwise.}
\end{cases}    
\end{align}

Finally, we sum $R^{\text{pos}}$ and $R^{\text{pun}}$ to obtain $R^{\text{acc}} = R^{\text{pos}} + R^{\text{pun}}$.

\subsection{The Learning Algorithm}
\label{sec:learning}

We employ GRPO as the reinforcement learning algorithm for STAR-R1, since GRPO samples and scores a group of responses based on the reward function, then calculates each response's relative advantage within the group. Compared to traditional Proximal Policy Optimization (PPO), GRPO eliminates the need for training an additional critic model by estimating the baseline from group scores, thereby significantly improving training efficiency.

Formally, given an initial image $I^{\text{initial}}$, a final image $I^{\text{final}}$, and a question description $Q$, the model first generates a group of $G$ responses $\{T_{1}, T_{2}, \ldots, T_{G}\}$. STAR-R1 then computes the format reward $R_{g}^{\text{format}}$ and accuracy reward $R_{g}^{\text{acc}}$ for each response, and sums them to obtain a set of rewards $\{R_{1}, R_{2}, \ldots, R_{G}\}$, where
$R_{g} = R_{g}^{\text{format}} + R_{g}^{\text{acc}}$. The advantage $\hat{A}_{g}$ for each response is then calculated by normalizing the rewards with respect to the group statistics: $\hat{A}_{g} = \frac{R_{g} - \mu_{G}}{\sigma_{G}}$, where $\mu_{G}$ and $\sigma_{G}$ are the mean and  the standard deviation of the rewards in the group.

Finally, based on the computed advantages and a KL penalty term that constrains the divergence between the trained policy and the reference policy, we obtain the following training objective, which is maximized to optimize the policy model:
\begin{align}
\nonumber
\mathcal{J}_{\text{GRPO}}(\theta) &= \mathbb{E}_{q \sim P(Q), \{T_g\}_{g=1}^G \sim \pi_{\theta_{old}}(T|q)} \\
&\quad \frac{1}{G} \sum_{g=1}^G \frac{1}{|o_i|} \left[ \min \left( R_{g} \hat{A}_{g}, \text{clip}\left( R_{g}, 1-\epsilon, 1+\epsilon \right) \hat{A}_{g} \right) - \beta D_{KL}[\pi_\theta \| \pi_{ref}] \right],
\end{align}
where $r_{g} = \frac{\pi_\theta(T_{g}|q)}{\pi_{\theta_{old}}(T_{g}|q)}$ is the probability ratio for observation $T_{g}$ under the current policy $\pi_\theta$ relative to the old policy $\pi_{\theta_{old}}$, $\pi_{ref}$ denotes the reference policy, and $\beta$ is a hyperparameter controlling the strength of the KL penalty.

\section{Experiments}
\label{sec:exp}
\textbf{Dataset.} As a challenging multimodal reasoning task, TVR requires the model to infer and describe single-step or multi-step transformations (i.e., changes in object attributes) between given initial and final images. To better reflect real-world scenarios, in addition to simpler tasks where both initial and final images share the same central viewpoint, TVR also includes more challenging cases where the model must match corresponding objects across different viewpoints (e.g., initial image from the central view while final image from the left/right view) and identify the transformations. 
After cleaning and filtering the original TRANCE dataset~\citep{hong2021transformation}, we randomly sample 9k training examples with both initial and final images from the central view. From the remaining data, we randomly sample 4.5k test examples comprising initial images from the central view, paired with final images from left, center, and right views. We provide further details in Appendix A.

\textbf{Implementations and Baselines.} We employ Qwen2.5-VL-7B as the base model for our main experiments, and conduct all experiments on a single computation node equipped with 8×H20 GPUs. Unlike existing approaches~\citep{feng2025video, meng2025mm} that require multi-stage training, our STAR-R1 demonstrates excellent reasoning capability through direct RL training without any supervised fine-tuning (SFT) as prerequisite. Due to current computational constraints, all RL experiments in this paper (unless otherwise specified) are trained for 2 epochs. To comprehensively evaluate the reasoning capabilities of STAR-R1, we conduct extensive comparisons with numerous existing models, including both closed-source (e.g., Qwen-VL~\citep{bai2023qwenvl, wang2024qwen2vl, bai2025qwen25vl}, InterVL~\citep{chen2024internvl, chen2024internvl1.5, team2024internvl2}) and open-source (e.g., GPT-4o~\citep{hurst2024gpt}, Gemini\cite{team2024gemini}) alternatives.

To obtain a comprehensive, intuitive, and in-depth understanding of the model's reasoning capabilities, we employ multiple evaluation metrics. We first execute all transformations from the model's response on the initial image state, obtaining the predicted final state $S^{\text{final}}$. By comparing this with the ground truth final state $\hat{S}^{\text{final}}$, we compute the following two categories of metrics:

\begin{itemize}[left=1.0em, itemsep=2pt, parsep=2pt, topsep=0pt]
    \item \textbf{Sample-level Metrics}. (1) \textit{Attribute Accuracy}: measuring whether each attribute (color, shape, size, material) of all objects is predicted correctly. (2) \textit{Total Accuracy (TAcc)}: measuring whether all four attributes of all transformations are predicted correctly. (3) \textit{Attribute Difference (Diff)}: counting the number of differing attributes between $S^{\text{final}}$ and $\hat{S}^{\text{final}}$. (4) \textit{Normalized Difference (Norm Diff)}: dividing Diff by the total number of transformations, eliminating scale effect.

    \item \textbf{Population-level Metrics}. To evaluate how object number influences  complexity and challenge, we categorize samples into four groups based on object count: $\{1$-$3, 4$-$6, 7$-$8, 9$-$10\}$. For each group, we calculate \textit{Object Total Accuracy}: the total accuracy within each category.

\end{itemize}

\subsection{Main results}

As shown in ~\cref{tab:main_results}, our model STAR-R1 achieves state-of-the-art performance across all eleven evaluation metrics when compared with advanced MLLMs. This demonstrates both the exceptional capability of STAR-R1 and the remarkable potential of the reinforcement learning paradigm for visual reasoning tasks. Specifically, we have the following observations: (1) For all metrics, STAR-R1 shows significant improvement over the base model Qwen-2.5-VL-7B. It substantially outperforms not only  InternVL2 with comparable model scale, but also larger-scale models like QWen-2.5-VL-7B. Remarkably, in terms of \textit{TAcc}, STAR-R1 surpasses the closed-source commercial models GPT-4o and Gemini-1.5-pro by \textbf{37.9\%} and \textbf{45.5\%}, respectively. (2) Compared to the SFT paradigm, our pure RL approach without any cold-start data achieves approximately \textbf{13\%} performance gain of \textit{TAcc}, providing strong evidence that RL can significantly enhance model capabilities for complex visual reasoning scenarios. (3) For the four \textit{Attribute Accuracy} metrics, we observe comparable performance across all attributes. This indicates robust stability in our learning framework without exhibiting catastrophic forgetting or attribute dominance. (4) For the four population-level metrics, model performance shows a steep decline as the number of objects increases. This trend reflects the compounding challenges in complex scenes, where difficulties accumulate across the entire reasoning pipeline from object identification to object association and finally to change perception.

\input{Full_Results}

\subsection{Anthropomorphic Behavior: RL VS SFT}
\label{sec:rlvssft}
In this section, we conduct a thorough comparison between SFT and RL training paradigms by evaluating their performance on both In-Domain (ID) and Out-Of-Domain (OOD) test samples, where ID samples contain initial and final images from the same center viewpoint while OOD samples contain images from different viewpoints (center-to-left/right). The OOD setting presents a greater challenge, as models  must distinguish genuine object transformations from viewpoint-induced variations.
As shown in ~\cref{tab:sft_rl}, SFT achieves better accuracy on ID data, but RL demonstrates remarkable superiority in the OOD setting, outperforming SFT by \textbf{23\%}, highlighting its exceptional generalization capability for reasoning in unseen views.

\input{SFT_RL}

\begin{figure*}[!t]
    \centering    
    \includegraphics[width=\textwidth]{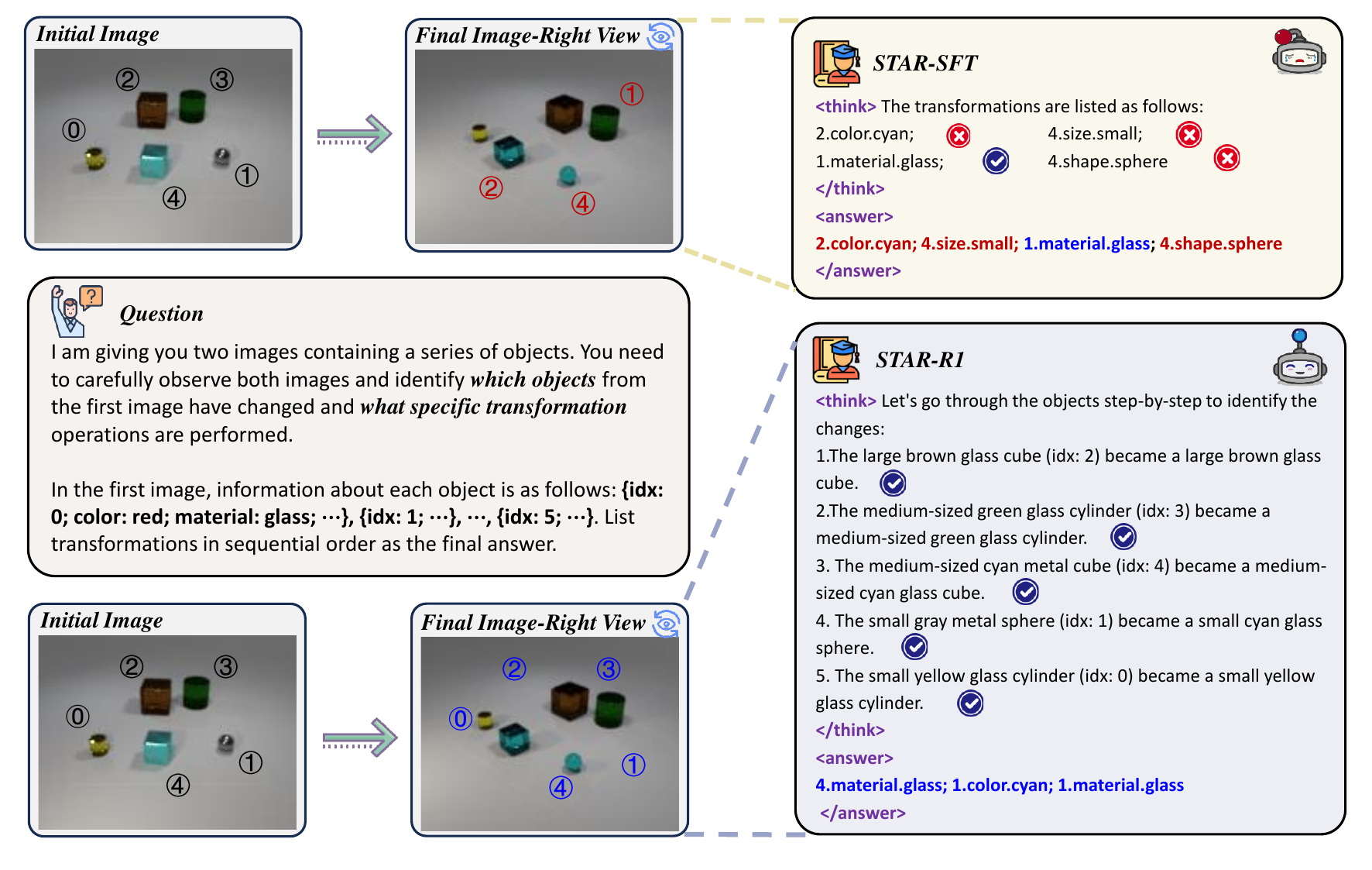} 
    \caption{A case study comparing the reasoning processes of STAR-SFT and STAR-R1. STAR-SFT performs only cursory comparisons, failing to detect view changes, which results in incorrect object matching and answers. In contrast, STAR-R1 systematically compares all objects step-by-step, identifies the view shift, and accurately tracks object correspondences, producing correct results. \textcolor{blue}{Blue} indicates correct answers while \textcolor{red}{red} denotes incorrect ones.}
    \label{fig:case_study}
    \vspace{-15pt}
\end{figure*}

Further case studies reveal that RL model's superior performance in the OOD scenario stems fundamentally from its capability to systematically compare each object between initial and final images. As illustrated in ~\cref{fig:case_study}, the STAR-SFT model typically performs cursory comparisons of only a few objects. This approach suffices for ID scenarios with consistent viewpoints, where differences are often immediately apparent. 
However, in OOD scenarios, viewpoint transitions necessitate cross-verification through multiple object comparisons to reliably identify transformations, mirroring human reasoning processes. For example, STAR-SFT misidentifies objects 1, 4, and 3 in the final image as objects 4, 2, and 1, respectively. This suggests that the model fails to effectively account for viewpoint-induced object displacement, erroneously retaining the object IDs from the initial viewpoint in the transformed viewpoint, thus leading to incorrect object correspondence and a significant degradation in prediction accuracy. In contrast, STAR-R1 meticulously examine all objects, including untransformed ones. In multi-object scenes, this comprehensive comparison enables precise cross-view mapping by referring to untransformed objects, thereby preventing performance degradation from mismatches and significantly improving robustness to viewpoint changes. Importantly, during the whole process, our prompts do not include \textbf{any} information related to view invariance/variation. Quantitatively, while processing consistent-viewpoint scenes (ID), RL model performs full object comparisons in 67\% of cases, but this ratio increases to \textbf{81\%} for viewpoint-shifting scenarios (OOD). This phenomenon indicates that the model, like humans, adapts its verification intensity based on situational complexity, performing more exhaustive comparisons when viewpoint changes demand higher confidence in object correspondences. Additional case studies are provided in Appendix D for further reference. Furthermore, we apply RL training to the STAR-SFT model and observe a series of intriguing experimental results and phenomena, with the findings presented in Appendix C.

\subsection{Training Curve of Response Length}

\begin{wrapfigure}[13]{r}[0pt]{0.48\columnwidth}
  \centering

  \includegraphics[width=\linewidth]{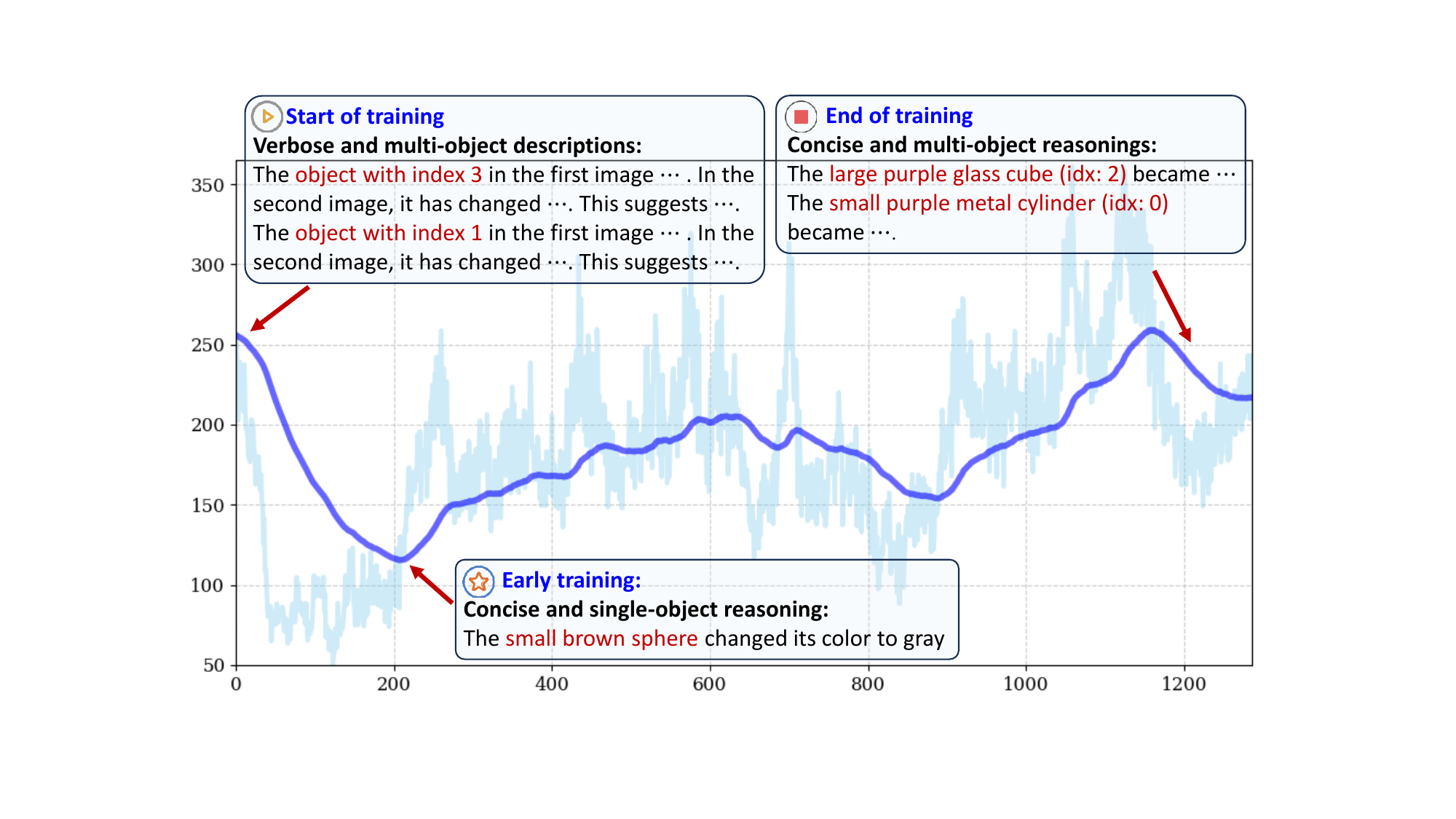}

  \caption{Dynamics of model response length. Zoom in to examine specific examples.}
  \label{fig:response_length}

\end{wrapfigure}

We analyze the training curve of the response length to further reveal the underlying insights. For additional details regarding the training curves, please refer to Appendix B. In ~\cref{fig:response_length}, the model's response length initially undergoes a rapid decline, followed by a gradual increase before eventually stabilizing. We observe that during the early phase, up until the point where the response length reaches its minimum, the model actively explores various reasoning strategies while progressively refining its language. 
For instance, the model's output transitions from verbose and multi-object descriptions, such as: "The object with index 3 in the first image is a gray cylinder of size 'large'. In the second image, it has changed to metallic gold. This suggests a transformation in color and material," to 
more concise and single-object reasoning, such as: "The small brown sphere changed its color to gray." 
At this stage, however, the model realizes that focusing on only a few objects may lead to incorrect mappings between objects across the two images or the omission of critical information, leaving room for further improvement in prediction accuracy. Consequently, while maintaining its refined reasoning style for individual objects, the model gradually explores a new and stable reasoning path, systematically comparing the states of all objects between the two images. This approach ultimately achieves high accuracy while keeping the reasoning length at an appropriate level.

\subsection{Ablation Studies}

In this section, to comprehensively investigate the impact of various factors on model performance, we conduct ablation studies on each component of the reward function, the volume of training data, the maximum response length limit, and the model size.

\textbf{Reward Design}
\label{sec:ab_reward}
We conducted systematic ablation studies on all components of the reward function in STAR-R1 to evaluate their contributions to model performance, with experimental results presented in \cref{table:ab_reward}. Our key observations are: (1) The model's performance degrades when either the object reward (Row 1) or attribute reward (Row 2) is removed, demonstrating that dense rewards facilitate more efficient exploration and improve answer accuracy. (2) Removing the punishment constraint for under-prediction of transformations (when $N_{pred}$ < $N_{gt}$) leads to significant accuracy deterioration (Row 3). This reveals that the punishment mechanism, which encourages exploration and prevents model inaction due to risk aversion, enables the model to search for better reasoning paths. (3) Eliminating the wrong-answer punishment causes the model to reward hacking by exhaustively enumerating all possible (object, attribute, value) triplets, leading to meaningless responses. To further validate the importance of the punishment design, we replace all punishment constraints with a unified negative absolute difference between predicted and ground-truth transformation counts: ($-|n- \hat{n}|$). As shown in Row 4, this yields markedly worse performance, confirming that timely punishment for incorrect answers is crucial for correcting the model's exploration path. We additionally conduct experiments with Naive GRPO (Row 5), which demonstrates limited suitability for TVR tasks, yielding merely subpar results.

\input{Ablation_reward_data}

\textbf{Training Data}
\label{sec:ab_data}
To evaluate the effect of training sizes, we randomly sample four subsets for training from the whole training set, containing 1000, 3000, 5000, and 7000 samples, respectively, and evaluate them on the original test set. The experimental results shown in ~\cref{table:ab_data} demonstrate that as the volume of training data increases, model performance improves continuously, but the growth rate of accuracy exhibits an initial increase followed by a subsequent decline. This suggests that insufficient data in the early stages constrains model learning, while expanded datasets facilitate a phase of rapid knowledge acquisition, leading to significantly accelerated performance gains.

\input{Ablation_size}

\textbf{Model Size}
\label{sec:ab_size}
To assess the influence of model size, we perform experiments by replacing the base model with Qwen-2.5VL-3B. The experimental results are presented in ~\cref{table:ab_size}. It is evident that the 3B model also exhibits a significant improvement in prediction accuracy after RL training, demonstrating the effectiveness of our method. However, under the same training settings, the performance gain of the 3B model is considerably smaller than that of the 7B model. This suggests that the capability of the base model plays a crucial role in the post-training phase, determining the model's ability to learn and adapt to new knowledge and scenarios. A stronger base model can achieve a higher upper bound of reasoning performance.

\section{Conclusion}
\label{sec:conclusion}
In this paper, we present STAR-R1, a pure RL-trained model equipped with our new reward design for the challenging spatial reasoning task---TVR. By integrating dense rewards for partial correctness with punishment rewards for penalizing shortcuts, STAR-R1 enables robust learning, achieving superior performance across all eleven evaluation metrics. Results reveal 
a mechanistic advantage of RL over SFT stems from comprehensively comparing objects in the scene, especially in cross-view out-of-domain scenarios. Further analysis uncovers STAR-R1's anthropomorphic behavior and non-linear training dynamic of response length, offering insights into RL’s capability in complex visual reasoning. This work validates the potential of the R1-Zero paradigm for advanced  reasoning tasks. Our findings may pave the way for future research in enhancing MLLMs’ spatial cognition, emphasizing RL’s capability to unlock human-like reasoning through structured exploration and fine-grained feedback.

\medskip

\bibliographystyle{plain}
\bibliography{Reference}

\appendix

\section{Dataset and Experimental Setup}
We first clean the TRANCE dataset~\citep{hong2021transformation} by removing samples containing redundant transformations. Then, we randomly sample 9,000 and 4,500 samples as the training set and test set, respectively, ensuring that the number of samples corresponding to each transformation length (ranging from 1 to 4) is equally distributed in both the training and test sets.

We employ Qwen2.5-VL-7B as our base model and utilize vLLM~\citep{kwon2023efficient} as the training framework. All experiments (including RL and SFT) are conducted on a single node equipped with 8 $\times$ H20 GPUs. Except for the training strategy mentioned in~\cref{app:sft_rl}, all other training runs adopt a single-stage procedure with 2 epochs, a batch size of 1 per GPU, and gradient accumulation over 2 steps, yielding a total of 1,286 training steps.

\section{Training Curves}

\begin{figure}[H]
    \centering    
    \includegraphics[width=\textwidth]{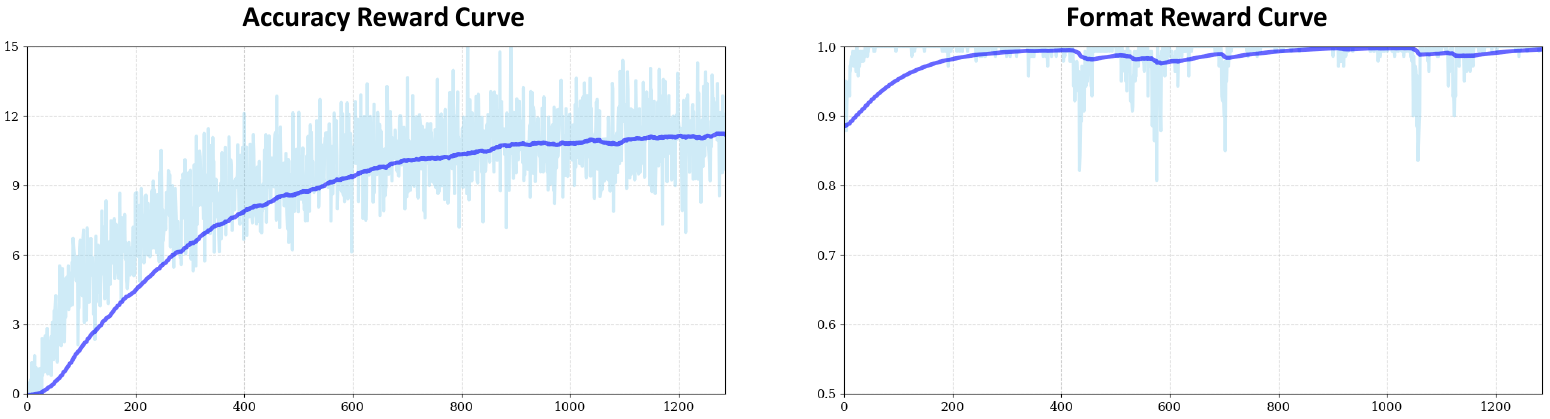} 
    \caption{Training curves about accuracy reward and format reward.}
    \label{fig:reward_curve}
    \vspace{-15pt}
\end{figure}

We illustrate the evolution of \textit{Total Accuracy} for STAR-RL and STAR-SFT models on OOD test data across training steps. The training dynamics plotted in ~\cref{fig:sft_rl_app} further show that SFT quickly plateaus around 1,200 steps on OOD data and eventually degrades with further training, whereas RL maintains steady performance growth throughout the training process, suggesting its unique ability to progressively unlock the model's reasoning capabilities for reasoning in complex scenarios.

Additionally, we plot further curves to analyze the training dynamics of STAR-R1. As shown in the right panel of ~\cref{fig:reward_curve}, we present the format reward curve of the STAR-R1. It can be observed that for models with strong foundational capabilities, such as Qwen-2.5-VL-7B, they already exhibit excellent instruction-following performance in the TVR task from the beginning, with most responses adhering to the required format. Thus, no additional customization for format reward is necessary. Moreover, in the left panel of ~\cref{fig:reward_curve}, we illustrate the accuracy reward curve. The model's accuracy reward steadily increases with training progression, ultimately stabilizing at a high value. This indicates that our designed reward function effectively encourages the model to engage in accurate, proactive, and efficient exploration, progressively unlocking the multimodal reasoning capabilities of the base model.

\begin{figure}[H]
    \centering    
    \includegraphics[width=\textwidth]{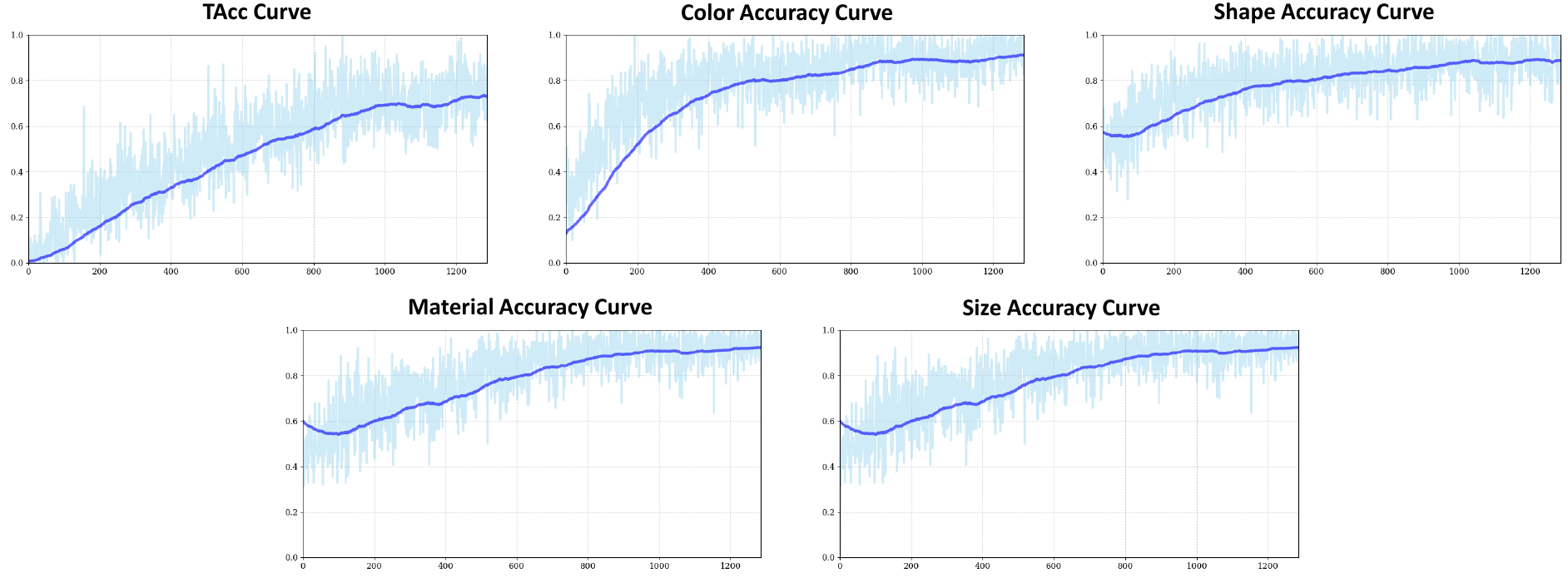} 
    \caption{Dynamics of the total accuracy as well as the
attribute accuracy for the four types throughout the training process.}
    \label{fig:acc_curve}
    \vspace{-15pt}
\end{figure}

As shown in ~\cref{fig:acc_curve}, these five curves illustrate the dynamics of the \textit{Total Accuracy} as well as the \textit{Attribute Accuracy} for the four types throughout the training process. All accuracy curves exhibit a steady increase during training and eventually stabilize at a high level of prediction accuracy. This demonstrates that our STAR-R1 effectively guides the model to efficiently explore the solution space and gradually develop stronger reasoning capabilities.

\input{SFT_RL_App}

\section{Training STAR-SFT with RL}
\label{app:sft_rl}
We additionally apply RL training to the SFT model. In Row 2 of ~\cref{tab:sft_rl_app}, results demonstrate that further training the model with RL not only achieves continuous improvement in ID performance but, more importantly, exhibits significant progress on OOD tasks. This indicates that RL successfully breaks through the performance bottleneck encountered during SFT training on OOD data. Furthermore, we observed that the two-phase training approach of STAR-SFT\&RL do not ultimately achieve the same performance on OOD data as pure RL training. We hypothesize that this may be due to imitation learning in the SFT phase locking the model's reasoning patterns, such that the subsequent RL training phase using the same amount of data can only marginally adjust the model's reasoning. This adjustment leads the model to attempt comparing more objects. 

To validate this conjecture, we further analyze the difference between the number of objects compared in each response and the total number of objects in the scene. The average difference for STAR-SFT is -3.3, while for STAR-SFT\&RL, it is -3.0. This indicates a slight increase in the number of objects compared during the reasoning process of the STAR-SFT\&RL. Consequently, this results in a modest improvement in OOD accuracy, though it still falls short of the performance achieved by the pure RL training.

\section{More Cases}
\begin{figure}[H]
    \centering    
    \includegraphics[width=\textwidth]{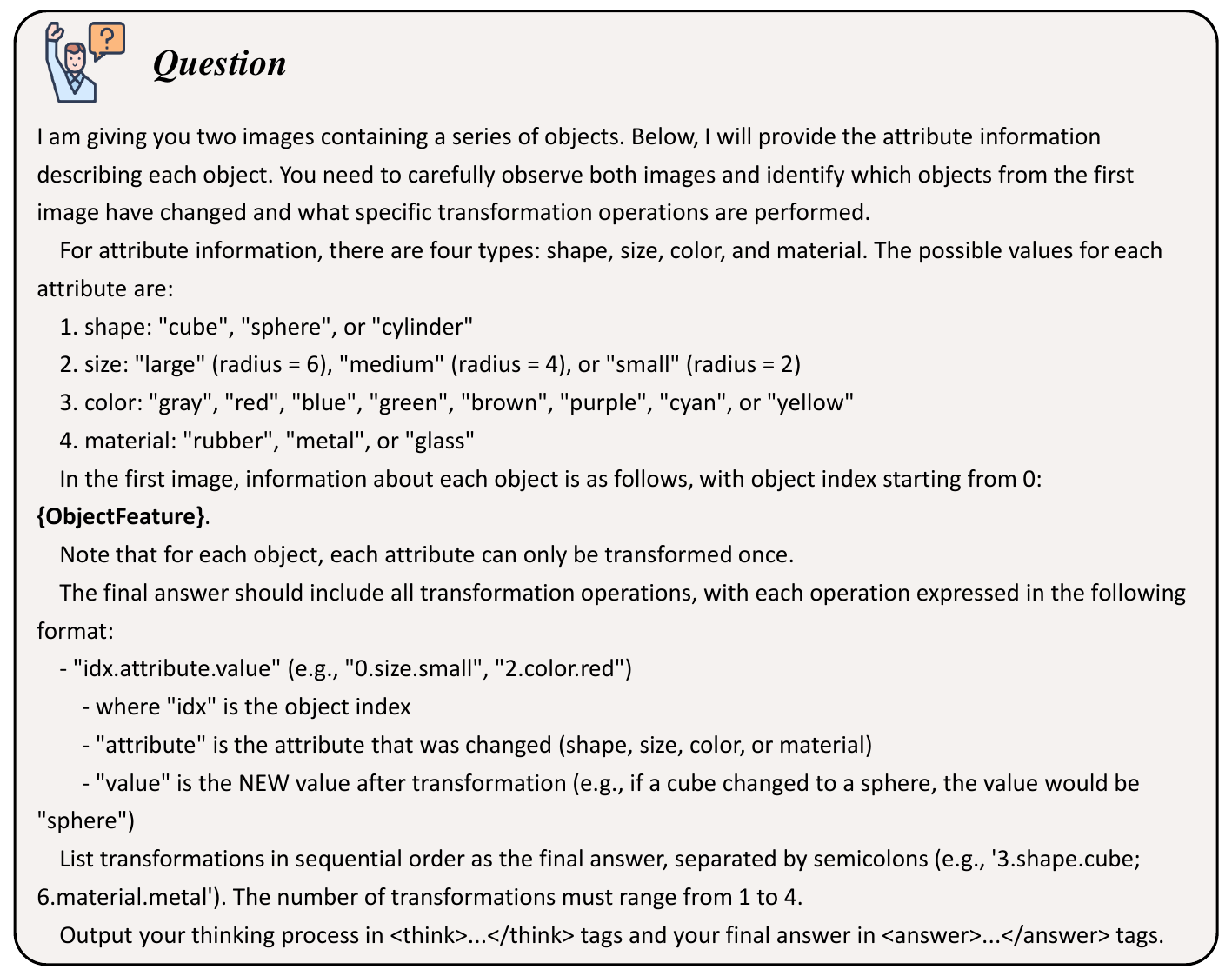} 
    \caption{The problem prompt.}
    \label{fig:prompt}
    \vspace{-15pt}
\end{figure}

In this section, we provide a detailed presentation of our problem prompt (shown in \cref{fig:prompt}) along with additional case studies. The \textbf{\{ObjectFeature\}} in \cref{fig:prompt} will be substituted by all features corresponding to each object in the initial image of every sample. An example is provided below: \{idx: 0; color: yellow; material: metal; shape: cylinder; size: medium\}, \{idx: 1; color: gray; material: rubber; shape: sphere; size: medium\}, \{idx: 2; color: blue; material: rubber; shape: cylinder; size: medium\}, \{idx: 3; color: yellow; material: metal; shape: sphere; size: medium\}, \{idx: 4; color: blue; material: rubber; shape: cube; size: large\}, \{idx: 5; color: brown; material: rubber; shape: sphere; size: medium\}, \{idx: 6; color: brown; material: rubber; shape: sphere; size: large\}.

\begin{figure}[H]
    \centering    
    \includegraphics[width=\textwidth]{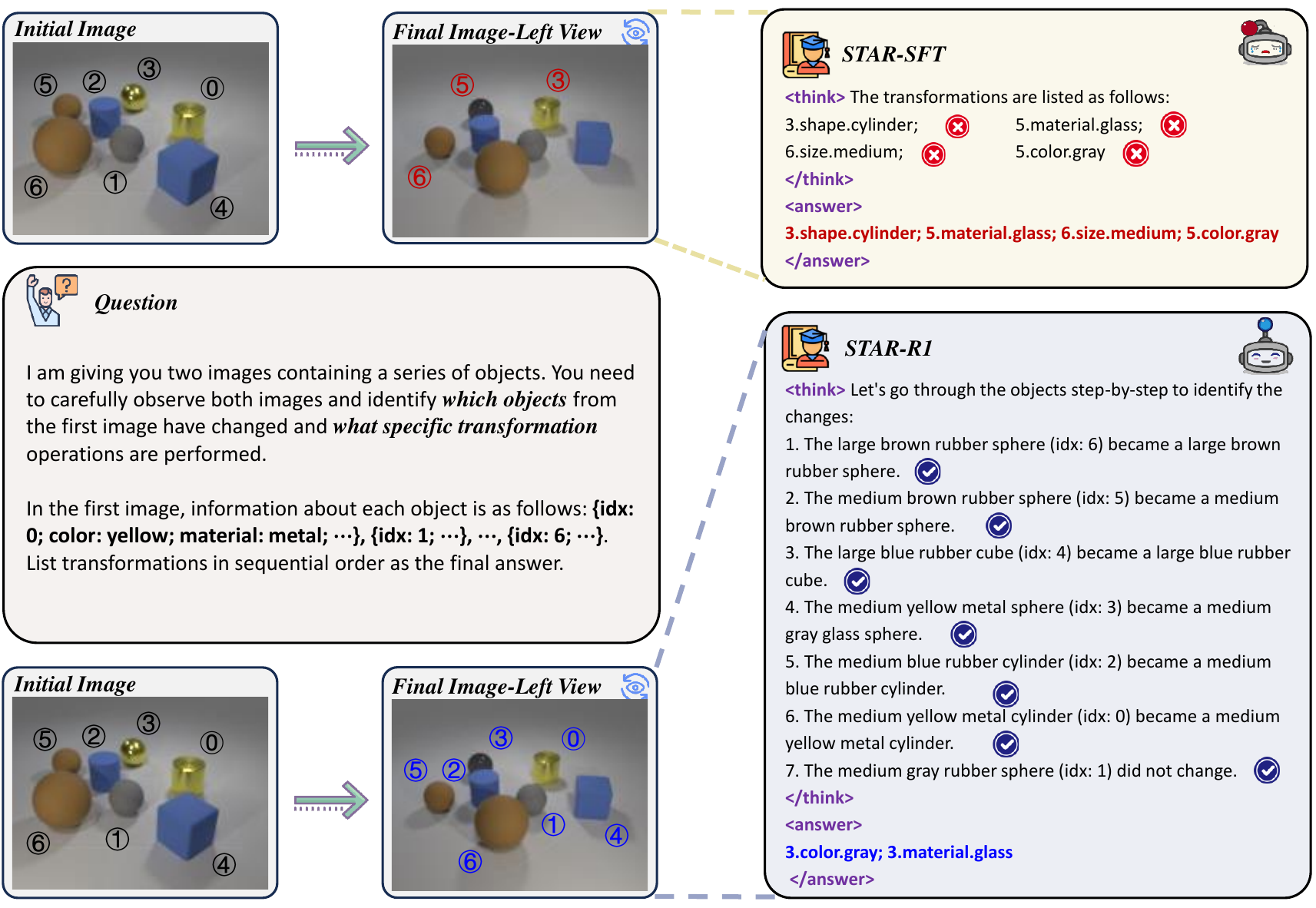} 
    \caption{A case study comparing the reasoning processes of STAR-SFT and STAR-R1. \textcolor{blue}{Blue} indicates correct answers while \textcolor{red}{red} denotes incorrect ones.}
    \label{fig:case_study6}
    \vspace{-15pt}
\end{figure}

\begin{figure}[H]
    \centering    
    \includegraphics[width=\textwidth]{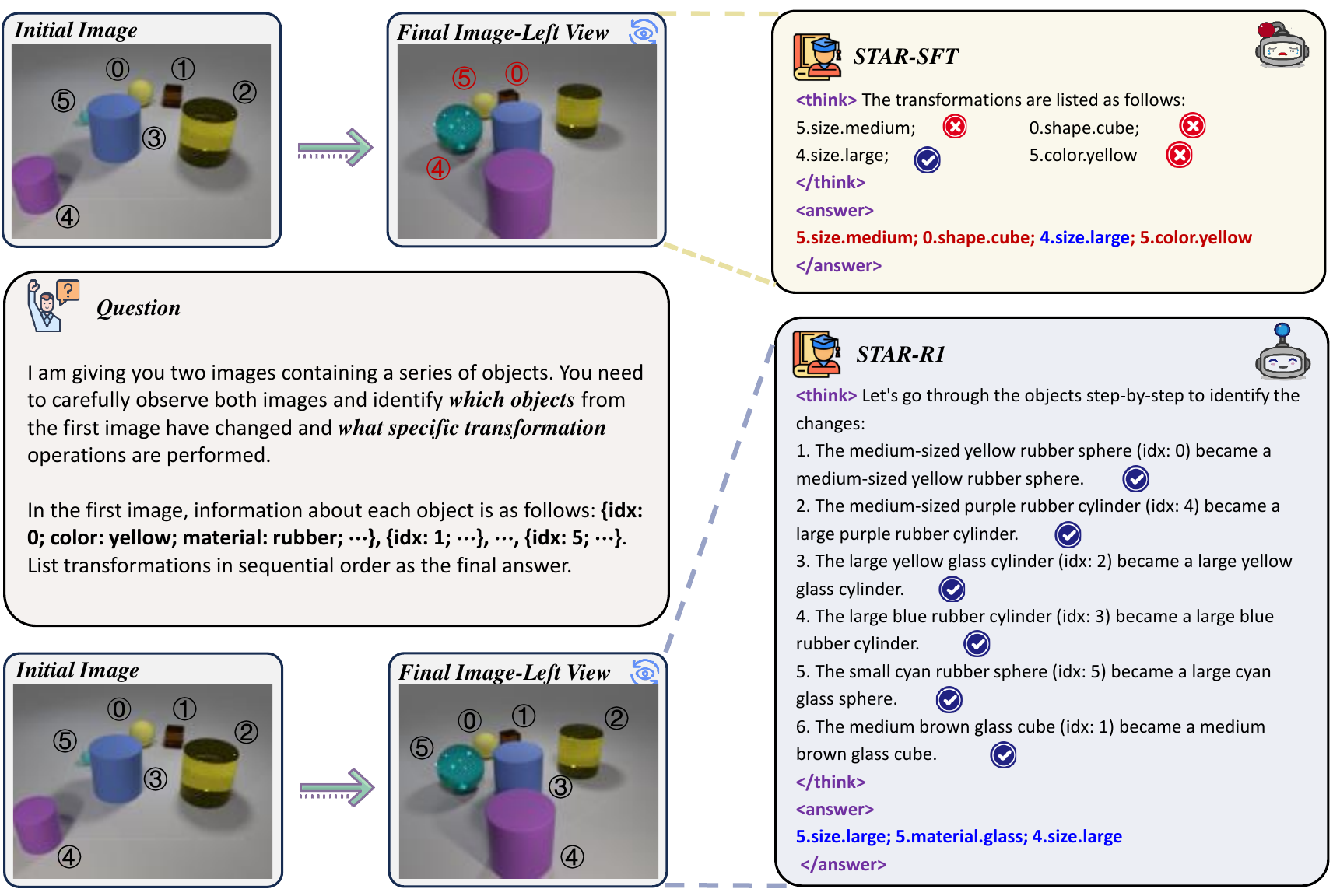} 
    \caption{A case study comparing the reasoning processes of STAR-SFT and STAR-R1. \textcolor{blue}{Blue} indicates correct answers while \textcolor{red}{red} denotes incorrect ones.}
    \label{fig:case_study7}
    \vspace{-15pt}
\end{figure}

\begin{figure}[H]
    \centering    
    \includegraphics[width=\textwidth]{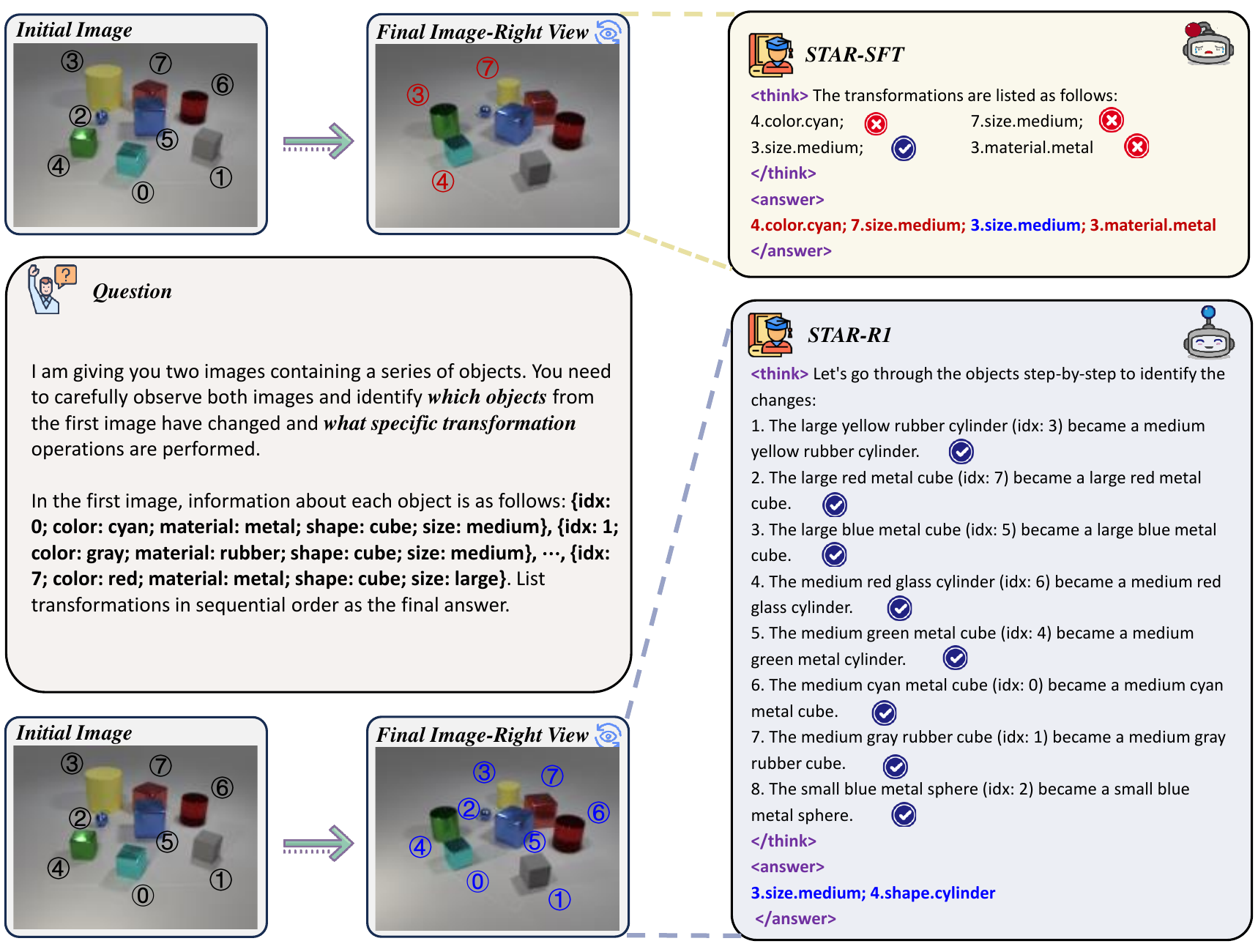} 
    \caption{A case study comparing the reasoning processes of STAR-SFT and STAR-R1. \textcolor{blue}{Blue} indicates correct answers while \textcolor{red}{red} denotes incorrect ones.}
    \label{fig:case_study8}
    \vspace{-15pt}
\end{figure}

The case studies shown in~\cref{fig:case_study6,fig:case_study7,fig:case_study8} further substantiate the argument presented in Section 4.2 of the main paper, namely that STAR-SFT merely engages in imitation learning and does not perceive changes in point of view. It erroneously associates objects located at the same positions in the initial and final images (referring to their positions within the images, not their positions in the real-world scene) as the same object, thereby significantly impairing the model's reasoning performance. 

In contrast, STAR-R1 demonstrates anthropomorphic behavior by systematically comparing the states of all objects between the initial and final images during its reasoning process to ascertain their correspondence. Consequently, without any explicit mention of viewpoint changes in the problem prompts, STAR-R1 successfully detects the perspective shift and achieves a significantly higher answer accuracy than STAR-SFT.

Since the STAR-SFT model fails to perceive changes in viewpoint, it mistakenly assumes that objects at the same position in two images are the same object. In ~\cref{fig:case_study6}, the STAR-SFT model incorrectly identifies objects numbered 0, 3, and 5 as objects numbered 3, 5, and 6. In ~\cref{fig:case_study7}, it misidentifies objects numbered 1, 0, and 5 as objects numbered 0, 5, and 4. In ~\cref{fig:case_study8}, the model confuses objects numbered 3, 4, and 0 with objects numbered 7, 3, and 4. In contrast, by comparing the states of all objects in both images, STAR-R1 correctly matches the same objects across the two images and ultimately outputs the correct answer.

\section{Limitations}
This work studies the problem of concurrent object transformations at a single timestamp. In contrast, real-world scenarios often involve temporally extended sequences of transformations with latent dependencies. Extending our approach to such long-term settings, accounting for temporal causality and state memory, is a key direction for future research.

\medskip

\end{document}

%% file: Full_Results.tex
\begin{table}[]
\caption{Results on TVR task. \textbf{Bold} indicates the best results.}
\label{tab:main_results}
\resizebox{\textwidth}{!}{
\begin{tabular}{lcccccccccccc}
\toprule
\multicolumn{1}{l}{\multirow{2}{*}{Model}} & \multicolumn{1}{c}{\multirow{2}{*}{Param}} & \multicolumn{1}{c}{\multirow{2}{*}{\textit{TAcc}$\uparrow$}} & \multicolumn{1}{c}{\multirow{2}{*}{\textit{Diff}$\downarrow$}} & \multicolumn{1}{c}{\multirow{2}{*}{\textit{NDiff}$\downarrow$}} & \multicolumn{4}{c}{Attribute Accuracy} & \multicolumn{4}{c}{Transformation Number} \\ \cmidrule(l){6-9} \cmidrule(l){10-13} 
\multicolumn{1}{c}{} & \multicolumn{1}{c}{} & \multicolumn{1}{c}{} & \multicolumn{1}{c}{} & \multicolumn{1}{c}{} & \multicolumn{1}{c}{\textit{Color}$\uparrow$} & \multicolumn{1}{c}{\textit{Shape}$\uparrow$} & \multicolumn{1}{c}{\textit{Size}$\uparrow$} & \multicolumn{1}{c}{\textit{Metarial}$\uparrow$} & \multicolumn{1}{c}{\textit{Num3}$\uparrow$} & \multicolumn{1}{c}{\textit{Num6}$\uparrow$} & \multicolumn{1}{c}{\textit{Num8}$\uparrow$} & \multicolumn{1}{c}{\textit{Num10}$\uparrow$} \\ \midrule
\rowcolor{gray!10}
\multicolumn{13}{c}{\textbf{\textit{Closed-Source Models}}} \\ \midrule
GPT-4o \cite{hurst2024gpt} & - & 23.5 & 1.96 & 0.76 & 61.7 & 69.3 & 62.2 & 61.0 & 49.3 & 27.3 & 14.6 & 9.6\\
\addlinespace[0.2em]
Gemini-2.0-flash \cite{team2024gemini} & - & 20.9 & 1.97 & 0.79 & 63.3 & 73.3 & 59.8 & 58.0  & 42.4 & 24.5 & 16.8 & 6.0 \\
\addlinespace[0.2em]
Gemini-1.5-Pro \cite{team2024gemini} & - & 15.9 & 2.46 & 1.00 & 54.4 & 72.2 & 54.6 & 48.9 & 37.4 & 18.5 & 8.6 & 4.7 \\ \midrule
\rowcolor{gray!10}
\multicolumn{13}{c}{\textbf{\textit{Open-Source Models}}} \\ \midrule
Deepseek-VL \cite{lu2024deepseek}  & 7B & 0.9 & 5.02 & 2.23 & 18.8 & 38.6 & 36.4 & 33.2 & 4.1 & 0.1 & 0.1 & 0.0 \\
Phi3.5V \cite{abdin2024phi} & 7B & 0.9 & 4.91 & 2.15 & 20.1 & 37.9 & 39.2 & 41.1 & 4.8 & 0.2 & 0.1 & 0.0 \\
\addlinespace[0.2em]
mPLUG-owl3 \cite{ye2024mplug} & 7B & 1.0 & 4.82 & 2.14 & 19.6 & 39.1 & 40.3 & 40.7 & 5.1 & 0.1 & 0.1 & 0.0 \\
\addlinespace[0.2em]
MiniCPM-V2.6 \cite{yao2024minicpm-v} & 7B & 1.0 & 4.78 & 2.11 & 22.5 & 40.2 & 41.3 & 38.3 & 5.4 & 0.1 & 0.1 & 0.1 \\
\addlinespace[0.2em]
LLaVA-OneVision \cite{li2024llava-onevision} & 7B & 1.1 & 4.67 & 2.06 & 22.4 & 43.3 & 42.5 & 40.6 & 5.8 & 0.1 & 0.1 & 0.0 \\
\addlinespace[0.2em]
Pixtral \cite{agrawal2024pixtral} & 12B & 3.0 & 3.02 & 1.44 & 30.3 & 53.6 & 54.7 & 52.2 & 10.8 & 1.9 & 0.7 & 0.3 \\
\addlinespace[0.2em]
InternVL2.5 \cite{chen2024internvl1.5} & 8B & 3.3 & 2.80 & 1.14 & 33.6 & 59.3 & 56.0 & 51.0 & 11.2 & 2.7 & 0.9 & 0.4 \\
\addlinespace[0.2em]
InternVL3 \cite{zhu2025internvl3} & 7B &  3.7 & 2.62 & 1.08 & 35.3 & 58.2 & 58.5 & 54.7 & 13.3 & 2.6 & 0.9 & 0.4 \\
\addlinespace[0.2em]
Qwen2-VL \cite{wang2024qwen2vl} & 2B & 0.3 & 5.23 & 2.40 & 14.9 & 45.0 & 48.4 & 54.1 & 1.0 & 0.3 & 0.0 & 0.0 \\
\addlinespace[0.2em]
Qwen2-VL \cite{wang2024qwen2vl} & 7B & 1.2 & 4.11 & 1.84 & 23.8 & 45.6 & 45.3 & 42.0 & 6.2 & 0.1 & 0.1 & 0.1 \\
\addlinespace[0.2em]
Qwen2.5-VL \cite{bai2025qwen25vl} & 3B & 2.3 & 3.53 & 1.51 & 25.6 & 47.2 & 52.2 & 44.5 & 8.8 & 1.5 & 0.5 & 0.3 \\
\addlinespace[0.2em]
Qwen2.5-VL \cite{bai2025qwen25vl} & 7B & 3.8 & 3.41 & 1.46 & 26.2 & 56.5 & 52.8 & 45.9 & 14.5 & 2.5 & 0.9 & 0.3 \\
\addlinespace[0.2em]
Qwen2.5-VL \cite{bai2025qwen25vl} & 32B & 9.1 & 3.21 & 1.38 & 41.1 & 57.3 & 50.6 & 36.6 & 27.5 & 9.8 & 2.5 & 1.4 \\
\midrule
STAR-SFT & 7B & 48.7 & 1.17 & 0.57 & 75.5 & 72.1 & 80.2 & 79.3 & 82.4 & 53.3 & 37.9 & 30.0 \\
\addlinespace[0.2em]
\rowcolor{blue!7.5}
STAR-R1 & 7B & \textbf{61.4} & \textbf{0.77} & \textbf{0.31} & \textbf{81.3} & \textbf{83.2} & \textbf{86.1} & \textbf{85.5} & \textbf{91.0} & \textbf{70.7} & \textbf{54.2} & \textbf{37.5} \\ \bottomrule
\end{tabular}
}
\vskip -0.2in
\end{table}

%% file: SFT_RL.tex
\begin{wrapfigure}[8]{r}[0pt]{0.48\columnwidth} 
  \vspace{-0.45cm}
  \centering 

  \begin{minipage}{0.48\columnwidth}
    \centering 
    \captionof{table}{Results in In-Domain (ID) and Out-Of-Domain (OOD) scenarios.}
    \label{tab:sft_rl} 
    \vspace{-0.15cm}
    \setlength{\tabcolsep}{3pt} 
    \fontsize{7.8pt}{10.2pt}\selectfont
    \begin{tabular}{lcccccc} 
      \toprule 
      & \multicolumn{3}{c}{ID} & \multicolumn{3}{c}{OOD} \\ 
      \cmidrule(lr){2-4} \cmidrule(lr){5-7} 
      Method & {\textit{TAcc}} & {\textit{Diff}} & {\textit{NDiff}} & {\textit{TAcc}} & {\textit{Diff}} & {\textit{NDiff}} \\ 
      \midrule 
      STAR-SFT & \textbf{84.2} & \textbf{0.22} & \textbf{0.08} & 30.9 & 1.65 & 0.83 \\ 
      STAR-R1 & 76.3 & 0.38 & 0.14 & \textbf{53.9} & \textbf{0.96} & \textbf{0.39} \\ 
      \bottomrule 
    \end{tabular} 
  \end{minipage}

\end{wrapfigure}

%% file: Ablation_reward_data.tex
\begin{table}[t!]
  \centering 

  \begin{minipage}{0.48\columnwidth} 
    \centering 
    \caption{Ablation studies on reward design.} 
    \label{table:ab_reward} 

    \setlength{\tabcolsep}{3pt} 
    \begin{scriptsize} 
      \begin{tabular}{lccccccc}
        \toprule
        \multicolumn{1}{l}{\multirow{2}{*}{Component}} & \multicolumn{1}{c}{\multirow{2}{*}{\textit{TAcc}}} & \multicolumn{1}{c}{\multirow{2}{*}{\textit{Diff}}} & \multicolumn{1}{c}{\multirow{2}{*}{\textit{NDiff}}} & \multicolumn{4}{c}{Transformation Number} \\ \cmidrule(l){5-8} 
\multicolumn{1}{c}{} & \multicolumn{1}{c}{} & \multicolumn{1}{c}{} & \multicolumn{1}{c}{} & {\textit{Num3}} & \multicolumn{1}{c}{\textit{Num6}} & \multicolumn{1}{c}{\textit{Num8}} & \multicolumn{1}{c}{\textit{Num10}} \\
        \midrule
        w/o obj & 58.0 & 0.90 & 0.37 & 89.8 & 66.5 & 49.5 & 34.6 \\
        \addlinespace[0.3em]
        w/o attr & 56.8 & 0.96 & 0.40 & 87.4 & 64.6 & 49.3 & 34.1 \\
        \addlinespace[0.3em]
        w/o up & 58.2 & 0.97 & 0.41 & 90.2 & 68.5 & 49.6 & 33.0 \\
        \addlinespace[0.3em]
        w/o pun & 54.3 & 1.05 & 0.44 & 85.5 & 61.9 & 46.8 & 31.4 \\
        \addlinespace[0.3em]
        w naive grpo & 54.5 & 1.02 & 0.43 & 85.2 & 63.7 & 46.1 & 31.1 \\
        \addlinespace[0.3em]
        STAR-R1 & \textbf{61.4} & \textbf{0.77} & \textbf{0.31} & \textbf{91.0} & \textbf{70.7} & \textbf{54.2} & \textbf{37.5} \\
        \bottomrule
      \end{tabular}
    \end{scriptsize}
  \end{minipage}
  \hfill
  \begin{minipage}{0.48\columnwidth} 
    \centering 
    \caption{Ablation studies on data volume.} 
    \label{table:ab_data}

    \setlength{\tabcolsep}{3pt} 
    \begin{scriptsize} 
      \begin{tabular}{lccccccc}
        \toprule
        \multicolumn{1}{l}{\multirow{2}{*}{Volume}} & \multicolumn{1}{c}{\multirow{2}{*}{\textit{TAcc}}} & \multicolumn{1}{c}{\multirow{2}{*}{\textit{Diff}}} & \multicolumn{1}{c}{\multirow{2}{*}{\textit{NDiff}}} & \multicolumn{4}{c}{Transformation Number} \\ \cmidrule(l){5-8} 
\multicolumn{1}{c}{} & \multicolumn{1}{c}{} & \multicolumn{1}{c}{} & \multicolumn{1}{c}{} & {\textit{Num3}} & \multicolumn{1}{c}{\textit{Num6}} & \multicolumn{1}{c}{\textit{Num8}} & \multicolumn{1}{c}{\textit{Num10}} \\
        \midrule
        1,000 & 13.8 & 2.32 & 0.95 & 29.0 & 16.2 & 9.0 & 4.9 \\
        \addlinespace[0.3em]
        3,000 & 23.7 & 1.97 & 0.88 & 49.0 & 25.3 & 17.2 & 10.3 \\
        \addlinespace[0.3em]
        5,000 & 40.7 & 1.49 & 0.65 & 75.1 & 47.0 & 30.1 & 19.5 \\
        \addlinespace[0.3em]
        7,000 & 52.6 & 1.08 & 0.46 & 82.7 & 59.6 & 45.0 & 31.2 \\
        \addlinespace[0.3em]
        9,000 & \textbf{61.4} & \textbf{0.77} & \textbf{0.31} & \textbf{91.0} & \textbf{70.7} & \textbf{54.2} & \textbf{37.5} \\
        \bottomrule
      \end{tabular}
    \end{scriptsize}
  \end{minipage}
\end{table}

%% file: Ablation_size.tex
\begin{wrapfigure}[9]{r}[0pt]{0.48\columnwidth} 
 \vspace{-0.45cm}
  \centering 

  \captionof{table}{Ablation studies on model size.} 
  \label{table:ab_size} 

  \setlength{\tabcolsep}{1.5pt} 
  \begin{scriptsize} 
    \begin{tabular}{lccccccc}
      \toprule
      \multicolumn{1}{l}{\multirow{2}{*}{Method}} & \multicolumn{1}{c}{\multirow{2}{*}{\textit{TAcc}}} & \multicolumn{1}{c}{\multirow{2}{*}{\textit{Diff}}} & \multicolumn{1}{c}{\multirow{2}{*}{\textit{NDiff}}} & \multicolumn{4}{c}{Transformation Number} \\ \cmidrule(l){5-8}
\multicolumn{1}{c}{} & \multicolumn{1}{c}{} & \multicolumn{1}{c}{} & \multicolumn{1}{c}{} & {\textit{Num3}} & \multicolumn{1}{c}{\textit{Num6}} & \multicolumn{1}{c}{\textit{Num8}} & \multicolumn{1}{c}{\textit{Num10}} \\
      \midrule
      Qwen2.5-VL-3B & 2.3 & 3.53 & 1.51 & 8.8 & 1.5 & 0.5 & 0.3 \\
      \addlinespace[0.3em]
      STAR-R1-3B & 28.6 & 1.84 & 0.79 & 55.2 & 32.4 & 20.4 & 13.5 \\
      \midrule
      Qwen2.5-VL-7B & 3.8 & 3.41 & 1.46 & 14.5 & 2.5 & 0.9 & 0.3 \\
      \addlinespace[0.3em]
      STAR-R1-7B & \textbf{61.4} & \textbf{0.77} & \textbf{0.31} & \textbf{91.0} & \textbf{70.7} & \textbf{54.2} & \textbf{37.5} \\
      \bottomrule
    \end{tabular}
  \end{scriptsize}
\end{wrapfigure}

%% file: SFT_RL_App.tex
\begin{figure*}[htbp]
  \begin{minipage}[]{0.48\columnwidth} 
    \centering
    \includegraphics[width=\linewidth]{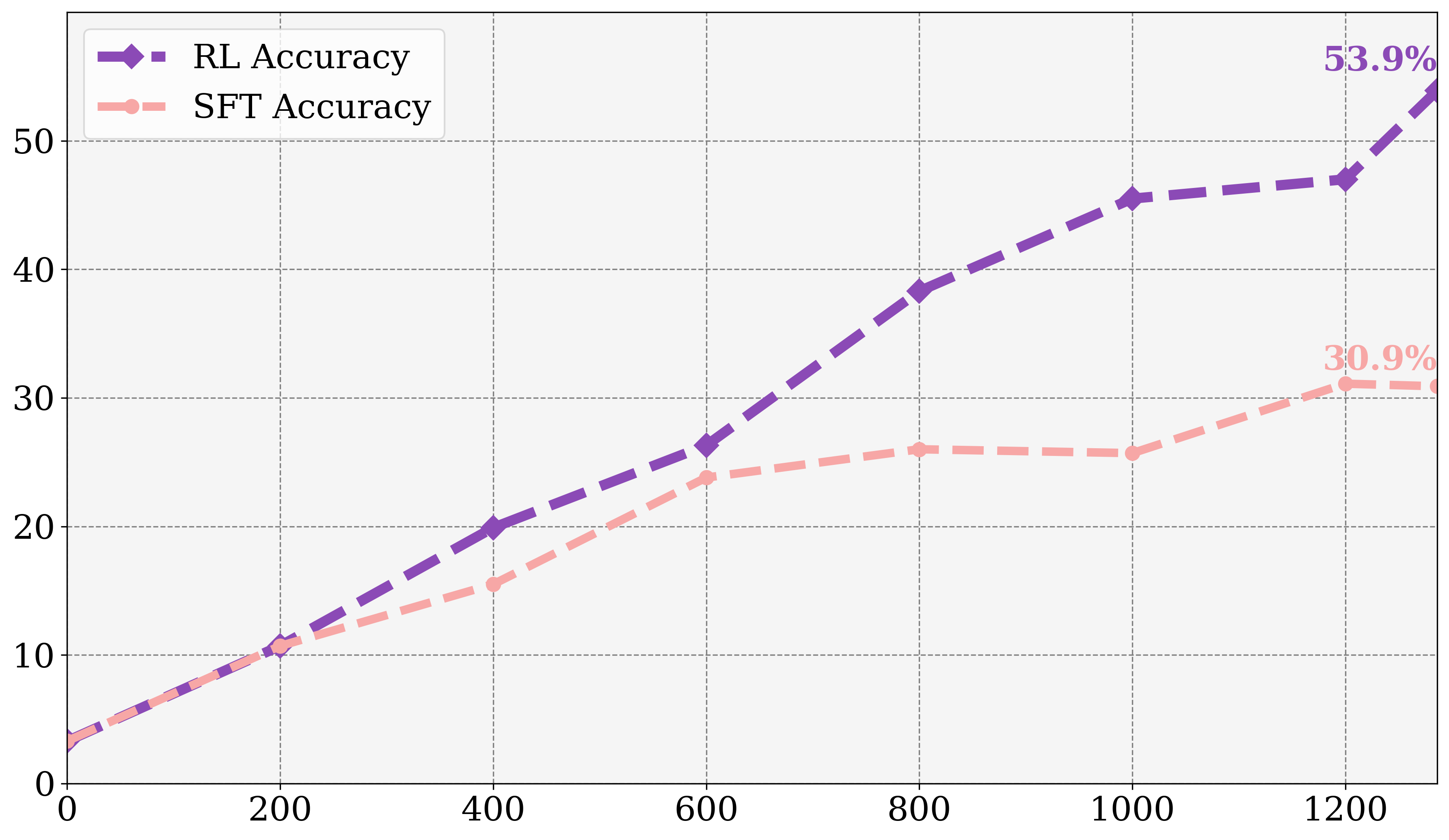}
    \caption{Accuracy curves of RL and SFT.}
    \label{fig:sft_rl_app}
  \end{minipage}
  \hfill
  \begin{minipage}[]{0.48\columnwidth} 
    \centering
    \captionof{table}{Comparison of RL vs. SFT performance on ID and OOD settings.}
    \label{tab:sft_rl_app}
    \setlength{\tabcolsep}{3pt} 
    \fontsize{8.0pt}{10.2pt}\selectfont
    \begin{tabular}{lcccccc} 
      \toprule 
      & \multicolumn{3}{c}{ID} & \multicolumn{3}{c}{OOD} \\ 
      \cmidrule(lr){2-4} \cmidrule(lr){5-7} 
      Method & TAcc & Diff & NDiff & TAcc & Diff & NDiff \\ 
      \midrule 
      STAR-SFT & 84.2 & 0.22 & 0.08 & 30.9 & 1.65 & 0.83 \\ 
      STAR-SFT\&RL & \textbf{87.7} & \textbf{0.19} & \textbf{0.07} & 36.5 & 1.53 & 0.71 \\ 
      STAR-R1 & 76.3 & 0.38 & 0.14 & \textbf{53.9} & \textbf{0.96} & \textbf{0.39} \\ 
      \bottomrule 
    \end{tabular} 
  \end{minipage}

\end{figure*}

%% file: neurips_2025.bbl
\begin{thebibliography}{10}

\bibitem{abdin2024phi}
Marah Abdin, Jyoti Aneja, Hany Awadalla, Ahmed Awadallah, Ammar~Ahmad Awan, Nguyen Bach, Amit Bahree, Arash Bakhtiari, Jianmin Bao, Harkirat Behl, et~al.
\newblock Phi-3 technical report: A highly capable language model locally on your phone.
\newblock {\em arXiv preprint arXiv:2404.14219}, 2024.

\bibitem{agrawal2024pixtral}
Pravesh Agrawal, Szymon Antoniak, Emma~Bou Hanna, Baptiste Bout, Devendra Chaplot, Jessica Chudnovsky, Diogo Costa, Baudouin De~Monicault, Saurabh Garg, Theophile Gervet, et~al.
\newblock Pixtral 12b.
\newblock {\em arXiv preprint arXiv:2410.07073}, 2024.

\bibitem{bai2023qwen}
Jinze Bai, Shuai Bai, Yunfei Chu, Zeyu Cui, Kai Dang, Xiaodong Deng, Yang Fan, Wenbin Ge, Yu~Han, Fei Huang, et~al.
\newblock Qwen technical report.
\newblock {\em arXiv preprint arXiv:2309.16609}, 2023.

\bibitem{bai2023qwenvl}
Jinze Bai, Shuai Bai, Shusheng Yang, Shijie Wang, Sinan Tan, Peng Wang, Junyang Lin, Chang Zhou, and Jingren Zhou.
\newblock Qwen-vl: A frontier large vision-language model with versatile abilities.
\newblock {\em arXiv preprint arXiv:2308.12966}, 1(2):3, 2023.

\bibitem{bai2025qwen25vl}
Shuai Bai, Keqin Chen, Xuejing Liu, Jialin Wang, Wenbin Ge, Sibo Song, Kai Dang, Peng Wang, Shijie Wang, Jun Tang, et~al.
\newblock Qwen2. 5-vl technical report.
\newblock {\em arXiv preprint arXiv:2502.13923}, 2025.

\bibitem{chen2025r1v}
Liang Chen, Lei Li, Haozhe Zhao, Yifan Song, and Vinci.
\newblock R1-v: Reinforcing super generalization ability in vision-language models with less than \$3.
\newblock \url{https://github.com/Deep-Agent/R1-V}, 2025.
\newblock Accessed: 2025-02-02.

\bibitem{chen2024we}
Lin Chen, Jinsong Li, Xiaoyi Dong, Pan Zhang, Yuhang Zang, Zehui Chen, Haodong Duan, Jiaqi Wang, Yu~Qiao, Dahua Lin, et~al.
\newblock Are we on the right way for evaluating large vision-language models?
\newblock {\em arXiv preprint arXiv:2403.20330}, 2024.

\bibitem{chen2025janus}
Xiaokang Chen, Zhiyu Wu, Xingchao Liu, Zizheng Pan, Wen Liu, Zhenda Xie, Xingkai Yu, and Chong Ruan.
\newblock Janus-pro: Unified multimodal understanding and generation with data and model scaling.
\newblock {\em arXiv preprint arXiv:2501.17811}, 2025.

\bibitem{chen2024expanding2.5}
Zhe Chen, Weiyun Wang, Yue Cao, Yangzhou Liu, Zhangwei Gao, Erfei Cui, Jinguo Zhu, Shenglong Ye, Hao Tian, Zhaoyang Liu, et~al.
\newblock Expanding performance boundaries of open-source multimodal models with model, data, and test-time scaling.
\newblock {\em arXiv preprint arXiv:2412.05271}, 2024.

\bibitem{chen2024internvl1.5}
Zhe Chen, Weiyun Wang, Hao Tian, Shenglong Ye, Zhangwei Gao, Erfei Cui, Wenwen Tong, Kongzhi Hu, Jiapeng Luo, Zheng Ma, et~al.
\newblock How far are we to gpt-4v? closing the gap to commercial multimodal models with open-source suites.
\newblock {\em Science China Information Sciences}, 67(12):220101, 2024.

\bibitem{chen2024internvl}
Zhe Chen, Jiannan Wu, Wenhai Wang, Weijie Su, Guo Chen, Sen Xing, Muyan Zhong, Qinglong Zhang, Xizhou Zhu, Lewei Lu, et~al.
\newblock Internvl: Scaling up vision foundation models and aligning for generic visual-linguistic tasks.
\newblock In {\em Proceedings of the IEEE/CVF conference on computer vision and pattern recognition}, pages 24185--24198, 2024.

\bibitem{cobbe2021training}
Karl Cobbe, Vineet Kosaraju, Mohammad Bavarian, Mark Chen, Heewoo Jun, Lukasz Kaiser, Matthias Plappert, Jerry Tworek, Jacob Hilton, Reiichiro Nakano, et~al.
\newblock Training verifiers to solve math word problems.
\newblock {\em arXiv preprint arXiv:2110.14168}, 2021.

\bibitem{fan2025sophiavl}
Kaixuan Fan, Kaituo Feng, Haoming Lyu, Dongzhan Zhou, and Xiangyu Yue.
\newblock Sophiavl-r1: Reinforcing mllms reasoning with thinking reward.
\newblock {\em arXiv preprint arXiv:2505.17018}, 2025.

\bibitem{feng2025video}
Kaituo Feng, Kaixiong Gong, Bohao Li, Zonghao Guo, Yibing Wang, Tianshuo Peng, Benyou Wang, and Xiangyu Yue.
\newblock Video-r1: Reinforcing video reasoning in mllms.
\newblock {\em arXiv preprint arXiv:2503.21776}, 2025.

\bibitem{fu2024video}
Chaoyou Fu, Yuhan Dai, Yongdong Luo, Lei Li, Shuhuai Ren, Renrui Zhang, Zihan Wang, Chenyu Zhou, Yunhang Shen, Mengdan Zhang, et~al.
\newblock Video-mme: The first-ever comprehensive evaluation benchmark of multi-modal llms in video analysis.
\newblock {\em arXiv preprint arXiv:2405.21075}, 2024.

\bibitem{fu2024mme}
Chaoyou Fu, Yi-Fan Zhang, Shukang Yin, Bo~Li, Xinyu Fang, Sirui Zhao, Haodong Duan, Xing Sun, Ziwei Liu, Liang Wang, et~al.
\newblock Mme-survey: A comprehensive survey on evaluation of multimodal llms.
\newblock {\em arXiv preprint arXiv:2411.15296}, 2024.

\bibitem{ghosh2023geneval}
Dhruba Ghosh, Hannaneh Hajishirzi, and Ludwig Schmidt.
\newblock Geneval: An object-focused framework for evaluating text-to-image alignment.
\newblock {\em Advances in Neural Information Processing Systems}, 36:52132--52152, 2023.

\bibitem{grattafiori2024llama3}
Aaron Grattafiori, Abhimanyu Dubey, Abhinav Jauhri, Abhinav Pandey, Abhishek Kadian, Ahmad Al-Dahle, Aiesha Letman, Akhil Mathur, Alan Schelten, Alex Vaughan, et~al.
\newblock The llama 3 herd of models.
\newblock {\em arXiv preprint arXiv:2407.21783}, 2024.

\bibitem{guo2025deepseek}
Daya Guo, Dejian Yang, Haowei Zhang, Junxiao Song, Ruoyu Zhang, Runxin Xu, Qihao Zhu, Shirong Ma, Peiyi Wang, Xiao Bi, et~al.
\newblock Deepseek-r1: Incentivizing reasoning capability in llms via reinforcement learning.
\newblock {\em arXiv preprint arXiv:2501.12948}, 2025.

\bibitem{hendrycks2020measuring}
Dan Hendrycks, Collin Burns, Steven Basart, Andy Zou, Mantas Mazeika, Dawn Song, and Jacob Steinhardt.
\newblock Measuring massive multitask language understanding.
\newblock {\em arXiv preprint arXiv:2009.03300}, 2020.

\bibitem{hendrycks2021measuring}
Dan Hendrycks, Collin Burns, Saurav Kadavath, Akul Arora, Steven Basart, Eric Tang, Dawn Song, and Jacob Steinhardt.
\newblock Measuring mathematical problem solving with the math dataset.
\newblock {\em arXiv preprint arXiv:2103.03874}, 2021.

\bibitem{hong2021transformation}
Xin Hong, Yanyan Lan, Liang Pang, Jiafeng Guo, and Xueqi Cheng.
\newblock Transformation driven visual reasoning.
\newblock In {\em Proceedings of the IEEE/CVF Conference on computer vision and pattern recognition}, pages 6903--6912, 2021.

\bibitem{hu2024ella}
Xiwei Hu, Rui Wang, Yixiao Fang, Bin Fu, Pei Cheng, and Gang Yu.
\newblock Ella: Equip diffusion models with llm for enhanced semantic alignment.
\newblock {\em arXiv preprint arXiv:2403.05135}, 2024.

\bibitem{hurst2024gpt}
Aaron Hurst, Adam Lerer, Adam~P Goucher, Adam Perelman, Aditya Ramesh, Aidan Clark, AJ~Ostrow, Akila Welihinda, Alan Hayes, Alec Radford, et~al.
\newblock Gpt-4o system card.
\newblock {\em arXiv preprint arXiv:2410.21276}, 2024.

\bibitem{jia2025omnispatial}
Mengdi Jia, Zekun Qi, Shaochen Zhang, Wenyao Zhang, Xinqiang Yu, Jiawei He, He~Wang, and Li~Yi.
\newblock Omnispatial: Towards comprehensive spatial reasoning benchmark for vision language models.
\newblock {\em arXiv preprint arXiv:2506.03135}, 2025.

\bibitem{kwon2023efficient}
Woosuk Kwon, Zhuohan Li, Siyuan Zhuang, Ying Sheng, Lianmin Zheng, Cody~Hao Yu, Joseph~E. Gonzalez, Hao Zhang, and Ion Stoica.
\newblock Efficient memory management for large language model serving with pagedattention.
\newblock In {\em Proceedings of the ACM SIGOPS 29th Symposium on Operating Systems Principles}, 2023.

\bibitem{li2024llava-onevision}
Bo~Li, Yuanhan Zhang, Dong Guo, Renrui Zhang, Feng Li, Hao Zhang, Kaichen Zhang, Peiyuan Zhang, Yanwei Li, Ziwei Liu, et~al.
\newblock Llava-onevision: Easy visual task transfer.
\newblock {\em arXiv preprint arXiv:2408.03326}, 2024.

\bibitem{li2024llava-next}
Feng Li, Renrui Zhang, Hao Zhang, Yuanhan Zhang, Bo~Li, Wei Li, Zejun Ma, and Chunyuan Li.
\newblock Llava-next-interleave: Tackling multi-image, video, and 3d in large multimodal models.
\newblock {\em arXiv preprint arXiv:2407.07895}, 2024.

\bibitem{li2024mvbench}
Kunchang Li, Yali Wang, Yinan He, Yizhuo Li, Yi~Wang, Yi~Liu, Zun Wang, Jilan Xu, Guo Chen, Ping Luo, et~al.
\newblock Mvbench: A comprehensive multi-modal video understanding benchmark.
\newblock In {\em Proceedings of the IEEE/CVF Conference on Computer Vision and Pattern Recognition}, pages 22195--22206, 2024.

\bibitem{li2025temporal}
Yuetai Li, Zhangchen Xu, Fengqing Jiang, Bhaskar Ramasubramanian, Luyao Niu, Bill~Yuchen Lin, Xiang Yue, and Radha Poovendran.
\newblock Temporal sampling for forgotten reasoning in llms.
\newblock {\em arXiv preprint arXiv:2505.20196}, 2025.

\bibitem{liang2025swsselfawareweaknessdrivenproblem}
Xiao Liang, Zhong-Zhi Li, Yeyun Gong, Yang Wang, Hengyuan Zhang, Yelong Shen, Ying~Nian Wu, and Weizhu Chen.
\newblock Sws: Self-aware weakness-driven problem synthesis in reinforcement learning for llm reasoning, 2025.

\bibitem{liu2024llava-1.5}
Haotian Liu, Chunyuan Li, Yuheng Li, and Yong~Jae Lee.
\newblock Improved baselines with visual instruction tuning.
\newblock In {\em Proceedings of the IEEE/CVF Conference on Computer Vision and Pattern Recognition}, pages 26296--26306, 2024.

\bibitem{liu2023llava}
Haotian Liu, Chunyuan Li, Qingyang Wu, and Yong~Jae Lee.
\newblock Visual instruction tuning.
\newblock {\em Advances in neural information processing systems}, 36:34892--34916, 2023.

\bibitem{liu2025ir3d}
Parker Liu, Chenxin Li, Zhengxin Li, Yipeng Wu, Wuyang Li, Zhiqin Yang, Zhenyuan Zhang, Yunlong Lin, Sirui Han, and Brandon~Y Feng.
\newblock Ir3d-bench: Evaluating vision-language model scene understanding as agentic inverse rendering.
\newblock {\em arXiv preprint arXiv:2506.23329}, 2025.

\bibitem{lu2024deepseek}
Haoyu Lu, Wen Liu, Bo~Zhang, Bingxuan Wang, Kai Dong, Bo~Liu, Jingxiang Sun, Tongzheng Ren, Zhuoshu Li, Hao Yang, et~al.
\newblock Deepseek-vl: towards real-world vision-language understanding.
\newblock {\em arXiv preprint arXiv:2403.05525}, 2024.

\bibitem{mangalam2023egoschema}
Karttikeya Mangalam, Raiymbek Akshulakov, and Jitendra Malik.
\newblock Egoschema: A diagnostic benchmark for very long-form video language understanding.
\newblock {\em Advances in Neural Information Processing Systems}, 36:46212--46244, 2023.

\bibitem{mathew2021docvqa}
Minesh Mathew, Dimosthenis Karatzas, and CV~Jawahar.
\newblock Docvqa: A dataset for vqa on document images.
\newblock In {\em Proceedings of the IEEE/CVF winter conference on applications of computer vision}, pages 2200--2209, 2021.

\bibitem{mayer2025ivispar}
Julius Mayer, Mohamad Ballout, Serwan Jassim, Farbod~Nosrat Nezami, and Elia Bruni.
\newblock ivispar--an interactive visual-spatial reasoning benchmark for vlms.
\newblock {\em arXiv preprint arXiv:2502.03214}, 2025.

\bibitem{meng2025mm}
F~Meng, L~Du, Z~Liu, Z~Zhou, Q~Lu, D~Fu, B~Shi, W~Wang, J~He, K~Zhang, et~al.
\newblock Mm-eureka: Exploring visual aha moment with rule-based large-scale reinforcement learning.
\newblock {\em arXiv preprint arXiv:2503.07365}, 2025.

\bibitem{peng2025lmm}
Yingzhe Peng, Gongrui Zhang, Miaosen Zhang, Zhiyuan You, Jie Liu, Qipeng Zhu, Kai Yang, Xingzhong Xu, Xin Geng, and Xu~Yang.
\newblock Lmm-r1: Empowering 3b lmms with strong reasoning abilities through two-stage rule-based rl.
\newblock {\em arXiv preprint arXiv:2503.07536}, 2025.

\bibitem{radford2021learning}
Alec Radford, Jong~Wook Kim, Chris Hallacy, Aditya Ramesh, Gabriel Goh, Sandhini Agarwal, Girish Sastry, Amanda Askell, Pamela Mishkin, Jack Clark, et~al.
\newblock Learning transferable visual models from natural language supervision.
\newblock In {\em International conference on machine learning}, pages 8748--8763. PmLR, 2021.

\bibitem{shao2024deepseekmath}
Zhihong Shao, Peiyi Wang, Qihao Zhu, Runxin Xu, Junxiao Song, Xiao Bi, Haowei Zhang, Mingchuan Zhang, YK~Li, Y~Wu, et~al.
\newblock Deepseekmath: Pushing the limits of mathematical reasoning in open language models.
\newblock {\em arXiv preprint arXiv:2402.03300}, 2024.

\bibitem{shen2025vlm}
Haozhan Shen, Peng Liu, Jingcheng Li, Chunxin Fang, Yibo Ma, Jiajia Liao, Qiaoli Shen, Zilun Zhang, Kangjia Zhao, Qianqian Zhang, et~al.
\newblock Vlm-r1: A stable and generalizable r1-style large vision-language model.
\newblock {\em arXiv preprint arXiv:2504.07615}, 2025.

\bibitem{song2025maniplvmr1reinforcementlearningreasoning}
Zirui Song, Guangxian Ouyang, Mingzhe Li, Yuheng Ji, Chenxi Wang, Zixiang Xu, Zeyu Zhang, Xiaoqing Zhang, Qian Jiang, Zhenhao Chen, Zhongzhi Li, Rui Yan, and Xiuying Chen.
\newblock Maniplvm-r1: Reinforcement learning for reasoning in embodied manipulation with large vision-language models, 2025.

\bibitem{Emu2}
Quan Sun, Yufeng Cui, Xiaosong Zhang, Fan Zhang, Qiying Yu, Zhengxiong Luo, Yueze Wang, Yongming Rao, Jingjing Liu, Tiejun Huang, and Xinlong Wang.
\newblock Generative multimodal models are in-context learners.
\newblock 2023.

\bibitem{sun2023eva-clip}
Quan Sun, Yuxin Fang, Ledell Wu, Xinlong Wang, and Yue Cao.
\newblock Eva-clip: Improved training techniques for clip at scale.
\newblock {\em arXiv preprint arXiv:2303.15389}, 2023.

\bibitem{tang2025lego}
Kexian Tang, Junyao Gao, Yanhong Zeng, Haodong Duan, Yanan Sun, Zhening Xing, Wenran Liu, Kaifeng Lyu, and Kai Chen.
\newblock Lego-puzzles: How good are mllms at multi-step spatial reasoning?
\newblock {\em arXiv preprint arXiv:2503.19990}, 2025.

\bibitem{team2024gemini}
Gemini Team, Petko Georgiev, Ving~Ian Lei, Ryan Burnell, Libin Bai, Anmol Gulati, Garrett Tanzer, Damien Vincent, Zhufeng Pan, Shibo Wang, et~al.
\newblock Gemini 1.5: Unlocking multimodal understanding across millions of tokens of context.
\newblock {\em arXiv preprint arXiv:2403.05530}, 2024.

\bibitem{team2024internvl2}
OpenGVLab Team.
\newblock Internvl2: Better than the best—expanding performance boundaries of open-source multimodal models with the progressive scaling strategy, 2024.

\bibitem{touvron2023llama}
Hugo Touvron, Thibaut Lavril, Gautier Izacard, Xavier Martinet, Marie-Anne Lachaux, Timoth{\'e}e Lacroix, Baptiste Rozi{\`e}re, Naman Goyal, Eric Hambro, Faisal Azhar, et~al.
\newblock Llama: Open and efficient foundation language models.
\newblock {\em arXiv preprint arXiv:2302.13971}, 2023.

\bibitem{touvron2023llama2}
Hugo Touvron, Louis Martin, Kevin Stone, Peter Albert, Amjad Almahairi, Yasmine Babaei, Nikolay Bashlykov, Soumya Batra, Prajjwal Bhargava, Shruti Bhosale, et~al.
\newblock Llama 2: Open foundation and fine-tuned chat models.
\newblock {\em arXiv preprint arXiv:2307.09288}, 2023.

\bibitem{wang2025solidgeomeasuringmultimodalspatial}
Peijie Wang, Chao Yang, Zhong-Zhi Li, Fei Yin, Dekang Ran, Mi~Tian, Zhilong Ji, Jinfeng Bai, and Cheng-Lin Liu.
\newblock Solidgeo: Measuring multimodal spatial math reasoning in solid geometry, 2025.

\bibitem{wang2024qwen2vl}
Peng Wang, Shuai Bai, Sinan Tan, Shijie Wang, Zhihao Fan, Jinze Bai, Keqin Chen, Xuejing Liu, Jialin Wang, Wenbin Ge, et~al.
\newblock Qwen2-vl: Enhancing vision-language model's perception of the world at any resolution.
\newblock {\em arXiv preprint arXiv:2409.12191}, 2024.

\bibitem{wang2024lvbench}
Weihan Wang, Zehai He, Wenyi Hong, Yean Cheng, Xiaohan Zhang, Ji~Qi, Xiaotao Gu, Shiyu Huang, Bin Xu, Yuxiao Dong, et~al.
\newblock Lvbench: An extreme long video understanding benchmark.
\newblock {\em arXiv preprint arXiv:2406.08035}, 2024.

\bibitem{wang2024emu3}
Xinlong Wang, Xiaosong Zhang, Zhengxiong Luo, Quan Sun, Yufeng Cui, Jinsheng Wang, Fan Zhang, Yueze Wang, Zhen Li, Qiying Yu, et~al.
\newblock Emu3: Next-token prediction is all you need.
\newblock {\em arXiv preprint arXiv:2409.18869}, 2024.

\bibitem{wei2022chain}
Jason Wei, Xuezhi Wang, Dale Schuurmans, Maarten Bosma, Fei Xia, Ed~Chi, Quoc~V Le, Denny Zhou, et~al.
\newblock Chain-of-thought prompting elicits reasoning in large language models.
\newblock {\em Advances in neural information processing systems}, 35:24824--24837, 2022.

\bibitem{wei2023cmath}
Tianwen Wei, Jian Luan, Wei Liu, Shuang Dong, and Bin Wang.
\newblock Cmath: Can your language model pass chinese elementary school math test?
\newblock {\em arXiv preprint arXiv:2306.16636}, 2023.

\bibitem{xie2025logic}
Tian Xie, Zitian Gao, Qingnan Ren, Haoming Luo, Yuqian Hong, Bryan Dai, Joey Zhou, Kai Qiu, Zhirong Wu, and Chong Luo.
\newblock Logic-rl: Unleashing llm reasoning with rule-based reinforcement learning.
\newblock {\em arXiv preprint arXiv:2502.14768}, 2025.

\bibitem{xu2024llava}
Guowei Xu, Peng Jin, Li~Hao, Yibing Song, Lichao Sun, and Li~Yuan.
\newblock Llava-o1: Let vision language models reason step-by-step.
\newblock {\em arXiv preprint arXiv:2411.10440}, 2024.

\bibitem{xu2025redstar}
Haotian Xu, Xing Wu, Weinong Wang, Zhongzhi Li, Da~Zheng, Boyuan Chen, Yi~Hu, Shijia Kang, Jiaming Ji, Yingying Zhang, et~al.
\newblock Redstar: Does scaling long-cot data unlock better slow-reasoning systems?
\newblock {\em arXiv preprint arXiv:2501.11284}, 2025.

\bibitem{yang2024qwen2.5}
An~Yang, Baosong Yang, Beichen Zhang, Binyuan Hui, Bo~Zheng, Bowen Yu, Chengyuan Li, Dayiheng Liu, Fei Huang, Haoran Wei, et~al.
\newblock Qwen2. 5 technical report.
\newblock {\em arXiv preprint arXiv:2412.15115}, 2024.

\bibitem{yang2025deepcritic}
Wenkai Yang, Jingwen Chen, Yankai Lin, and Ji-Rong Wen.
\newblock Deepcritic: Deliberate critique with large language models.
\newblock {\em arXiv preprint arXiv:2505.00662}, 2025.

\bibitem{yang2025towards}
Wenkai Yang, Shuming Ma, Yankai Lin, and Furu Wei.
\newblock Towards thinking-optimal scaling of test-time compute for llm reasoning.
\newblock {\em arXiv preprint arXiv:2502.18080}, 2025.

\bibitem{yao2024minicpm}
Yuan Yao, Tianyu Yu, Ao~Zhang, Chongyi Wang, Junbo Cui, Hongji Zhu, Tianchi Cai, Haoyu Li, Weilin Zhao, Zhihui He, et~al.
\newblock Minicpm-v: A gpt-4v level mllm on your phone.
\newblock {\em arXiv preprint arXiv:2408.01800}, 2024.

\bibitem{yao2024minicpm-v}
Yuan Yao, Tianyu Yu, Ao~Zhang, Chongyi Wang, Junbo Cui, Hongji Zhu, Tianchi Cai, Haoyu Li, Weilin Zhao, Zhihui He, et~al.
\newblock Minicpm-v: A gpt-4v level mllm on your phone.
\newblock {\em arXiv preprint arXiv:2408.01800}, 2024.

\bibitem{ye2024mplug}
Jiabo Ye, Haiyang Xu, Haowei Liu, Anwen Hu, Ming Yan, Qi~Qian, Ji~Zhang, Fei Huang, and Jingren Zhou.
\newblock mplug-owl3: Towards long image-sequence understanding in multi-modal large language models.
\newblock {\em arXiv preprint arXiv:2408.04840}, 2024.

\bibitem{yu2023mm}
Weihao Yu, Zhengyuan Yang, Linjie Li, Jianfeng Wang, Kevin Lin, Zicheng Liu, Xinchao Wang, and Lijuan Wang.
\newblock Mm-vet: Evaluating large multimodal models for integrated capabilities.
\newblock {\em arXiv preprint arXiv:2308.02490}, 2023.

\bibitem{yue2024mmmu}
Xiang Yue, Yuansheng Ni, Kai Zhang, Tianyu Zheng, Ruoqi Liu, Ge~Zhang, Samuel Stevens, Dongfu Jiang, Weiming Ren, Yuxuan Sun, et~al.
\newblock Mmmu: A massive multi-discipline multimodal understanding and reasoning benchmark for expert agi.
\newblock In {\em Proceedings of the IEEE/CVF Conference on Computer Vision and Pattern Recognition}, pages 9556--9567, 2024.

\bibitem{zhong2023agieval}
Wanjun Zhong, Ruixiang Cui, Yiduo Guo, Yaobo Liang, Shuai Lu, Yanlin Wang, Amin Saied, Weizhu Chen, and Nan Duan.
\newblock Agieval: A human-centric benchmark for evaluating foundation models.
\newblock {\em arXiv preprint arXiv:2304.06364}, 2023.

\bibitem{zhou2024mlvu}
Junjie Zhou, Yan Shu, Bo~Zhao, Boya Wu, Shitao Xiao, Xi~Yang, Yongping Xiong, Bo~Zhang, Tiejun Huang, and Zheng Liu.
\newblock Mlvu: A comprehensive benchmark for multi-task long video understanding.
\newblock {\em arXiv preprint arXiv:2406.04264}, 2024.

\bibitem{zhu2025internvl3}
Jinguo Zhu, Weiyun Wang, Zhe Chen, Zhaoyang Liu, Shenglong Ye, Lixin Gu, Yuchen Duan, Hao Tian, Weijie Su, Jie Shao, et~al.
\newblock Internvl3: Exploring advanced training and test-time recipes for open-source multimodal models.
\newblock {\em arXiv preprint arXiv:2504.10479}, 2025.

\end{thebibliography}
